\documentclass{article} 
\usepackage{iclr2024_conference,times}

\usepackage{url}

\usepackage{amsmath}
\usepackage{graphicx}
\usepackage[colorlinks=true, allcolors=blue]{hyperref}

\usepackage{qiangstyle}

\newcommand{\lion}{Lion}
\newcommand{\adam}{Adam}
\newcommand{\adamw}{AdamW}

\newcommand{\rebuttal}[1]{{\color{black}{#1}}}

\usepackage{tabularx}

\title{
Lion Secretly Solves Constrained Optimization, As Lyapunov Predicts 
}

\iclrfinalcopy

\author{
Lizhang Chen\thanks{Equal Contribution} ~~~~~~~~~ 
Bo Liu\footnotemark[1] ~~~~~~~~~ 
Kaizhao Liang\footnotemark[1]  ~~~~~~~~~ 
Qiang Liu \\
The University of Texas at Austin\\
\texttt{\{lzchen,bliu,kaizhaol,lqiang\}@utexas.edu}
}

\begin{document}

\maketitle

\begin{abstract}
Lion (Evolved Sign Momentum), a new optimizer discovered through program search, has shown promising results in training large AI models. It performs comparably or favorably to AdamW but with greater memory efficiency. As we can expect from the results of a random search program, Lion incorporates elements from several existing algorithms, including signed momentum, decoupled weight decay, Polyak, and Nesterov momentum, but does not fit into any existing category of theoretically grounded optimizers. Thus, even though Lion appears to perform well as a general-purpose optimizer for a wide range of tasks, its theoretical basis remains uncertain. This lack of theoretical clarity limits opportunities to further enhance and expand Lion's efficacy.

This work aims to demystify Lion. Based on both continuous-time and discrete-time analysis, we demonstrate that Lion is a theoretically novel and principled approach for minimizing a general loss function $f(x)$ while enforcing a bound constraint $\norm{x}_\infty \leq 1/\lambda$. Lion achieves this through the incorporation of decoupled weight decay, where $\lambda$ represents the weight decay coefficient. Our analysis is made possible by the development of a new Lyapunov function for the Lion updates. It applies to a broader family of Lion-$\phi$ algorithms, where the $\text{sign}(\cdot)$ operator in Lion is replaced by the subgradient of a convex function $\phi$, leading to the solution of a general composite optimization problem of $\min_x f(x) + \phi^*(x)$. Our findings provide valuable insights into the dynamics of Lion and pave the way for further improvements and extensions of Lion-related algorithms.

\end{abstract}

\section{Introduction}
\label{sec::intro}

Optimization serves as the cornerstone in training contemporary AI models.
Given the immense computational demands associated with training large AI models,
the design of an effective optimizer emerges as a paramount endeavor.

Traditionally, efficient optimizers are devised by machine learning experts based on theoretical insights~\citep{bernstein_signsgd_2018, kingma2017adam, loshchilov2017decoupled, hazan_revisiting_2022}.
\adam{}~\citep{kingma2014adam} and its variant \adamw{}~\citep{loshchilov2017decoupled} remain the most widely employed methods in deep learning.
Recently, however, a new optimization named \lion{} (Evolved Sign Momentum)~\citep{chen2023symbolic} was discovered by an evolutionary search algorithm~\citep{real2020automl} applied to a symbolically represented program space~\citep{bello2017neural}. Lion has been shown to achieve at least comparable performance to AdamW on a wide range of tasks while reducing memory cost and training time~\citep{chen2023symbolic}.

However, as the outcome of a stochastic search algorithm, Lion does not have an \emph{a priori theoretical guarantee by design}. It is still uncertain whether Lion can be regarded as a reliable and legitimate general-purpose optimization algorithm, despite the reported positive results on a large, yet finite, set of tasks~\citep{chen2023symbolic}. The lack of theoretical understanding also significantly restricts the potential for improving and extending Lion to obtain better new optimizers.

In this work, we demonstrate that Lion, along with a broader family of Lion-$\phi$ algorithms, can be established as a theoretically novel and intriguing approach for solving optimization problems with convex regularization or constraints. This is surprising because Lion was discovered in a search space that includes arbitrary symbolic operations and was not designed with any theoretical guarantees. This discovery opens up promising opportunities for developing improved optimizers by leveraging the existing success of Lion.

\paragraph{Lion: Evolved Sign Momentum}  
The update rule of Lion for minimizing 
a loss 
$f(x)$ on $\RR^d$ is   
\bbb \label{equ:lion1} 
\text{Lion:}~~~~~~~~
\begin{split}
m_{t+1} & =  \btwo m_t  - (1-\btwo)\dd f(x_t),  \\ 
x_{t+1} & = x_t + \lr (\sign(\bone m_{t} - (1-\bone) \dd f(x_t) ) -  {\lambda x_t}), 
\end{split}
\eee 
where $m_t\in \RR^d$ is the momentum, $\lr>0$ is the learning rate,  $\bone, \btwo \in [0,1]$ are two momentum related coefficients, and $\lambda\geq 0$ is a weight decay coefficient. 
A default value of $\bone = 0.9$ and $\btwo = 0.99$ 
was suggested in \citet{chen2023symbolic}, 
with which the \lion{} update rule can be written directly as 
\bb \vv  x_{t+1} \gets (1-\epsilon \lambda )\vv x_t - \lr~ \sign\left ( (10+1){\vv g_{t}} + 0.99\vv g_{t-1} + 0.99^2\vv g_{t-2} + \cdots + 0.99^k \vv g_{t-k} + \cdots  \right ), \ee 
where $g_t = \dd f(x_t)$.  
Here the update of  $x_t$  
combines a weight decay term with coefficient $(1-\epsilon \lambda)$,
and the sign of a weighted average of the trajectory gradients. 
Notably, the weight of the current gradient $g_t$ is increased by $(\beta_2-\beta_1)/((1-\beta_2)\beta_1)\approx 10$ times 
compared with typical exponential moving average of 
gradients as used in the classical Polyak momentum \citep{polyak1964some}.  

One can think of Lion as made by ``splicing" the elements of 
many existing algorithms in Lion, which 
is exactly what an efficient search program can do when given a proper search space~\citep{peng2020pyglove,chen2023symbolic,bello2017neural}. 
The update of the momentum $m_t$ is common to
the Polyak momentum-based algorithms and yields the exponential moving average part of  the update.  
What sets it apart is the unique update of $x_t$, 
which uses the combination of three key elements: 

i)  \textbf{[Sign Reshaper]} The use of the $\sign(\cdot)$ function for update, similar to signed gradient descent and signed momentum \cite{bernstein2018signsgd, crawshaw_robustness_2022}, can be viewed 
as an extreme way of normalizing the magnitude 
of the coordinate-wise updates. 
It is closed related to normalized gradient~\citep[]{ 
levy2016power, 
murray2019revisiting}
and adaptive gradient methods such as Adam \cite{kingma2014adam} and RMSprop~\cite{tieleman2012lecture}. 
Note that Adam can be viewed as 
signed momentum 
with an adaptive variance based step size \citep{balles2018dissecting}, 
which might be the key factor explaining the gap between Adam and SGD \citep{kunstner2023noise}.

 ii) \textbf{[Gradient Enhancement]} 
When using $\beta_2 > \beta_1$,
the importance of the current gradient $g_t$
is increased
compared to the exponential moving average in
standard Polyak momentum update.
It can be shown that Polyak momentum with
this gradient enhancement results in Nesterov momentum,
and leads to the well-known acceleration phenomenon \citep[e.g.,][]{shi2021understanding}.

 iii) \textbf{[Decoupled Weight Decay]} The weight decay term $\lambda x_t$ 
 outside of the gradient and $\sign(\cdot)$. 
 Such idea of  the  \emph{decoupled} 
 weight decay is what make  AdamW \citep{loshchilov2019decoupled} significantly outperform the vanilla Adam in training large AI models. 


 As demonstrated by the empirical findings of \citet{chen2023symbolic} and subsequent research,
the combination of these elements has been shown to make Lion perform well on a wide range of problems, including image classification, language models, and diffusion models~\citep{chen2023symbolic}.

However, it remains unclear whether the combination of these elements yield a theoretically valid and convergent general-purpose optimizer. 
Furthermore, the use of decoupled weight decay adds to the uncertainty regarding what optimization problem Lion aims to solve: due to its interaction with other parts of the algorithm, decoupled weight decay is always not equivalent to simply introducing $\ell_2$ regularization \citep{loshchilov2017decoupled}.


\renewcommand{\arraystretch}{1.2} 
\begin{table}
\centering 
{\small 
\begin{tabular}{cc|c} \hline 
& Polyak Momentum  \citep{polyak1964some} & $\phi(x )= \norm{x}^2_2/2$, {$\gamma\lambda  =0$}, 
$\varepsilon = 0$ 
\\ \hline 
&Nesterov Momentum \citep{nesterov1983method} & $\phi(x) = \norm{x}^2_2/2$, {$\gamma \lambda =0$} \\ \hline 
&Signed Momentum \cite{bernstein2018signsgd} & $\phi(x) = \norm{x}_1^2$, $\varepsilon =0$, $\lambda =0$\\\hline 
&Hamiltonian Descent \citep{maddison2018hamiltonian} & $\varepsilon =0$, $\lambda =0$ \\ \hline 
&Hamiltonian Descent for Composite Objectives \citep{maddison2018hamiltonian} & $\varepsilon =0$, $\lambda >0$ \\ \hline 
&Dual Space Preconditioning \citep{maddison2021dual}, 
Mirror Descent \citep{nemirovskij1983problem}& $\varepsilon \gamma = 1$, $\lambda =0$  \\ \hline %
&Signed Gradient Descent \citep{bernstein2018signsgd}& $\phi(x)=\norm{x}_1$, $\varepsilon \gamma = 1$, $\lambda =0$ \\ \hline %
&Accelerated Mirror Descent \citep{krichene2015accelerated}& 
$\gamma =0$, $\varepsilon =0$, $\lambda > 0$  \\\hline 
& Frank–Wolfe \citep{frank1956algorithm} &  $\varepsilon\gamma=1$, $\lambda > 0$ \\\hline 
\end{tabular}
}
\caption{Lion-$\phi$ includes a large family of algorithms as special cases. See Section~\ref{sec:existing}}
\label{tab:lionk}
 \end{table}

\paragraph{``Lion King Meets Mr. Lyapunov"} 
We propose and analyze a general family of Lion-$\phi$ algorithms, 
in  which we replace the $\sign(\cdot)$ function in Lion 
with a subgradient $\dd\phi$ of a general convex function $\phi \colon \RR^d \to \RR$: 
\bbb \label{equ:lionK} 
\text{Lion-$\phi$:}~~~~~~~~
\begin{split}
m_{t+1} & =  \btwo m_t  - (1-\btwo)\dd f(x_t),  \\ 
x_{t+1} & = x_t + \lr (\dd\phi(\bone m_{t} - (1-\bone) \dd f(x_t) ) -  {\lambda x_t}). 
\end{split}
\eee 
Lion is recovered when $\phi(x) = \norm{x}_1$ and $\dd\phi(x) = \sign(x)$.  
Taking the continuous time limit of \eqref{equ:lionK}, we obtain 
the following ordinary differential equation (ODE): 
\bbb \label{equ:lionode}
\text{Lion-$\phi$ (ODE):}~~~~~~~~~~~~~
\begin{split}
& \dot m_t = - \alpha \dd f(x_t) - \gamma m_t \\ 
& \dot x_t =  \dd \phi(m_t - \varepsilon (\alpha \dd f(x_t) + \gamma m_t)) - \lambda x_t,
\end{split}
\eee  
Eq.~\eqref{equ:lionK} is the Euler discretization of Eq.~\eqref{equ:lionode} 
with step size $\epsilon$ in the case of $\alpha = \gamma$, with $\beta_1 = 1-\varepsilon \gamma$,  and $\beta_2 = 1-\epsilon \gamma$.
Lion-$\phi$ includes a broad set of algorithms as special cases,
as shown in Table~\ref{tab:lionk}. 

To avoid the complexities associated with regularity conditions, we can assume that  $\phi$ is continuously differentiable when discussing the ODE. But parallel results hold for the time discrete algorithm 
\eqref{equ:lionK} 
for general non-differentiable convex functions $\phi$. 

The crest of this work is to show that, 
when $\varepsilon \gamma \leq  1$,  
Lion-$\phi$ ODE 
solves the following optimization: 
\bbb \label{equ:opt_solve}
\min_{x\in\RR^d} F(x) \quad\text{with}\quad F(x)\defeq \alpha f(x) + \frac{\gamma }{\lambda }\phi^*(\lambda x), 
\eee 
where $\phi^* (x)\defeq \sup_z (x\tt z - \phi(z))$  is the conjugate function of $\phi$. 
Because we may have $\phi^*(x) = +\infty$ for some $x$,
 solving \eqref{equ:opt_solve} requires
 to enforce a constraint of $\lambda x \in \dom \phi^*$, where $\dom \phi^* \defeq \{x\colon \phi^*(x)<+\infty\}$ is the effective domain of $\phi^*.$
In the case of Lion, we have $\phi(x) = \norm{x}_1$ and hence $\phi^*(x) = \delta(\norm{x}_\infty \leq 1) $, where $\delta$ the $\infty$-indicator function with $\delta({\texttt{True}}) = 0$, $\delta(\texttt{False})=+\infty$. Hence,
Lion  solves the following bound-constrained optimization problem: 
\bbb \label{equ:lionboundc}
\min_{x\in\RR^d} f(x)~~~~s.t.~~~~ \norm{x}_\infty \leq 1/\lambda, 
\eee 
where the bound $1/\lambda$ is solely decided by the weight decay coefficient $\lambda$.

\begin{figure}[h!]
    \centering
    \includegraphics[width=.93\textwidth]{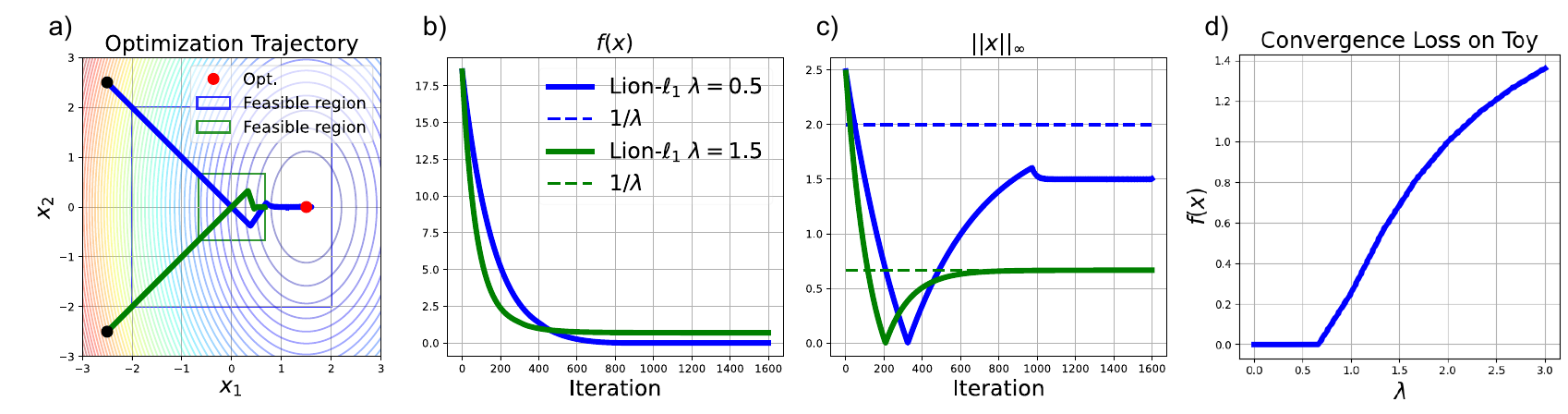}

    \vspace{0.1\baselineskip}
    \caption{
        (a)-(c)
        Trajectories of Lion  
        on 2D function $f(x) = (x_1 - 1.5)^2 + x_2^2$, 
        with $\lambda = 1.5$ and $\lambda = 0.5$ ((a)-(c)). 
        The boxes in a) represent the constraint set : blue box is for $\norm{x}_\infty \leq 1/\lambda$ with $\lambda = 0.5$, green box is for $\lambda = 1.5$.  
        \rebuttal{(d) $\lambda$ vs. the converged loss 
         We can see that the converged loss  starts to increase only when  $\lambda$ excel a threshold ($\lambda\geq 0.6$) to excluded the unconstrained minimum from the constrained set. }} 
    \label{fig:traj-1}
\end{figure}

\begin{figure}[h!]
    \centering
    \includegraphics[width=1\textwidth]{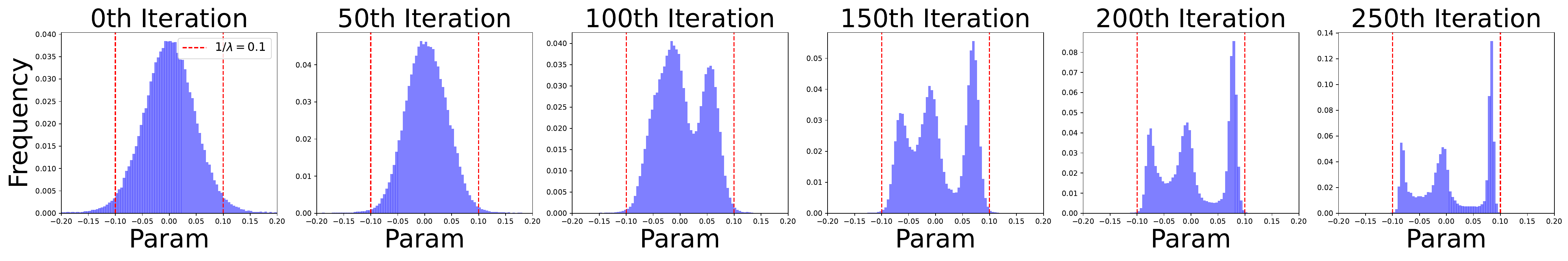}
    \vspace{-1\baselineskip}
    \caption{
        Histograms of the network parameters of ResNet-18 on CIFAR-10 trained by Lion with $\lambda = 10$.  The constraint of $\norm{x}_\infty\leq 1/\lambda$ (indicated by the red vertical lines) is satisfied within only $\sim$200 steps. 
    }  
    \label{fig:hist-row}
\end{figure}

\begin{figure}[h!]
    \centering
    \includegraphics[width=\textwidth]{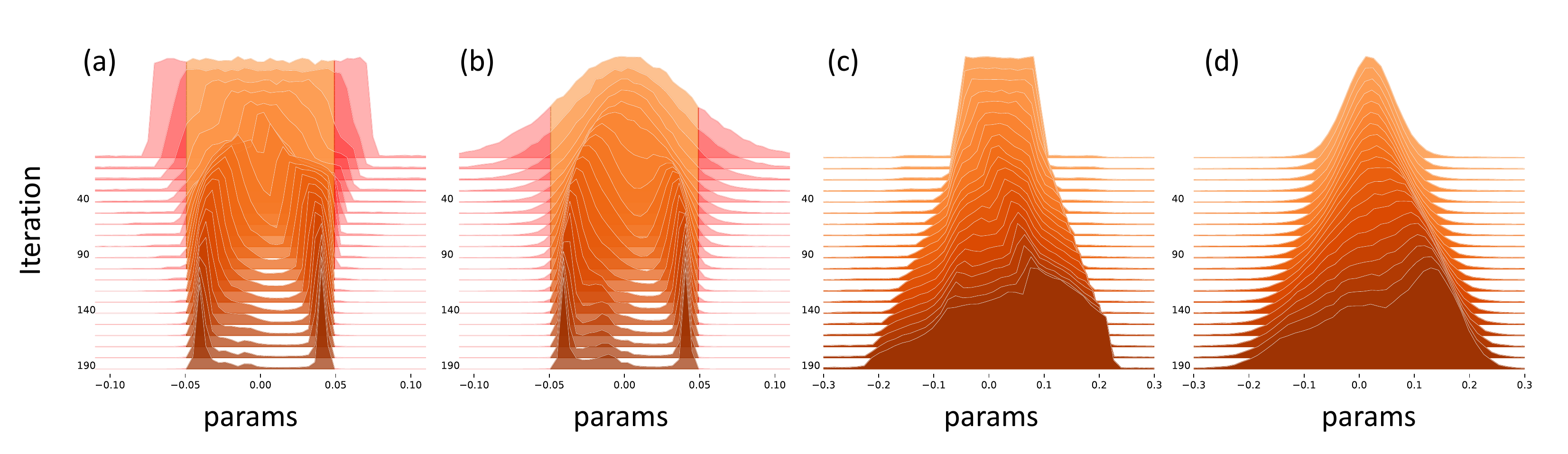}
    \vspace{-1\baselineskip}
    \caption{
     Evolution of histogram of parameter weights trained by Lion on ResNet-18 on CIFAR-10~\cite{he2016deep,krizhevsky2009learning}, with different $\lambda$ and initialization methods.  
        Frequency of network parameters in ResNet on the CIFAR-10 dataset across iterations. 
        {(a)}: Kaiming uniform initialization~\citep{he2015delving} and $\lambda = 20$. 
        {(b)}:  Kaiming normal initialization~\citep{he2015delving} and $\lambda = 20$. 
        {(c)}:  Kaiming uniform initialization~\citep{he2015delving} and $\lambda = 0$. 
        {(d)}:  Kaiming normal initialization~\citep{he2015delving} and $\lambda = 0$.
The weights are quickly confined into the bound $[-0.05, 0.05]$ with $\lambda =20$, 
while keep growing with zero weight decay ($\lambda =0$).
    }
    \label{fig:hist}
\end{figure}

Our proof shows that the Lion-$\phi$ dynamics 
consists of two phases: 

1) \textbf{[Phase 1]} When $\lambda x \not\in \dom \phi^*$, 
it exponentially decays the distance from $\lambda x_t$ to the set $\dom \phi^*$: 
$$
\dist(\lambda x_t, \dom \phi^*) \leq \exp(-\lambda (t-s) ) ~\dist(\lambda x_s, \dom\phi^*),~~~\forall s \leq t. 
$$
Hence, $\lambda x_t$ converges to $\dom \phi^*$ rapidly and stays within $\dom \phi^*$ once it arrived.  


2) \textbf{[Phase 2]} 
After $\lambda x_t$ enters $\dom \phi^*$, the dynamics minimizes the finite valued objective $F(x)$. This is proved by showing that the 
Lion-$\phi$ dynamics minimizes the following Lyapunov function: 
\bbb  \label{equ:H1}
H(x, m) = \alpha   f(x)   + 
\frac{ \gamma }{\lambda}
\phi^*(\lambda x) + 
\frac{1-\varepsilon \gamma }{1+ \varepsilon \lambda} 
(\phi^*(\lambda x) +  \phi(m)   - \lambda 
m\tt  x ). 
\eee
We show that, whenever $H(x_t,m_t)$ is finite, it is 
decreased monotonically (i.e., $\ddt H(x_t, m_t) \leq 0$) along trajectories of \eqref{equ:lionode} until a local minimum of point of $H(x,m)$ is reached.  

Furthermore,  we have  $F(x) = \min_{m} H(x, m)$,
and hence minimizing $H(x,m)$ is equivalent to minimizing $F(x)$; this is because the minimum of the  last term in \eqref{equ:H1} equals zero, 
$\min_m  \phi^*(\lambda x) +  \phi(m)   - \lambda 
m\tt  x =0 $, 
for any fixed $x$, by Fenchel-Young inequality.  

The discovery of this Lyapunov function is a new and non-trivial mathematical result.  But intuitively, 
one can see easily the connection of \eqref{equ:lionode} and \eqref{equ:opt_solve} by comparing their fixed points.  
Assume $\phi$ and $\phi^*$ are differentiable, 
then a fix point of \eqref{equ:lionode} must implies a stationary point of \eqref{equ:opt_solve}:  
\bb 
\underbrace{\alpha \dd f(x_t) + \gamma m_t = 0,~~~~ 
\dd\phi(m_t)=\lambda x_t}_{\text{fixed point of \eqref{equ:lionode}} } 
&&\implies && \underbrace{\alpha \dd f(x_t) + \gamma \dd\phi^*(\lambda x_t ) = 0,}_\text{{stationary point of \eqref{equ:opt_solve}}}
\ee 
where we used $\dd\phi(\dd\phi^*(x)) =x$, and $\dd_x \left (\frac{1}{\lambda} \phi^*(\lambda x) \right) = \dd\phi^*(\lambda x)$. 

\begin{figure}[h]
    \centering
    \includegraphics[width=0.31\textwidth]{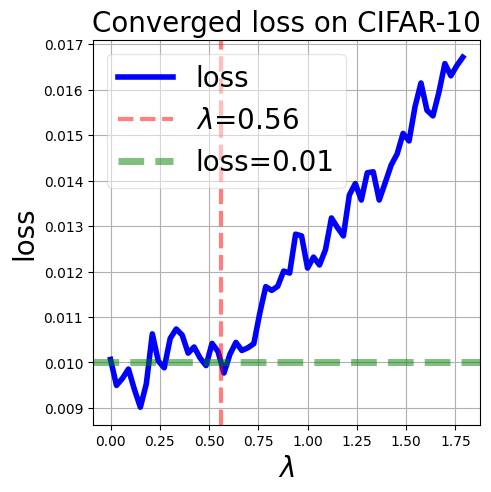}
    \includegraphics[width=0.32\textwidth]{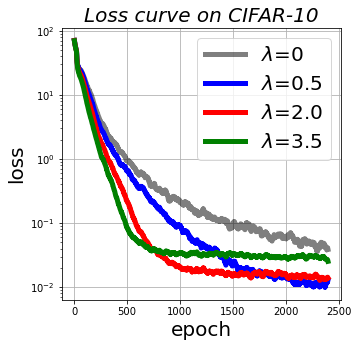}
    \caption{\rebuttal{Analysis of weight decay on CIFAR-10 using Lion. 
    a) The converged Loss vs. weight decay in Lion.
    We can see that the loss starts to increase only when $\lambda$ excel a threshold, which is expected from the constrained optimization view.
    b) The loss curves vs. epochs with different weight decays. 
    Larger weight decay $\lambda$ yields faster convergence (due to stronger Phase 1),  
    but may yield larger final loss when it is too large. 
    }}
    \label{fig:converged_loss}
\end{figure}

\rebuttal{
\paragraph{Why Should Lion Decay Weight?} 
From the analysis above, 
the role of weight decay $\lambda$ in Lion is two-fold: 

1) It alters the solution if $\lambda$ is large and 
the constraint $
\norm{x}_\infty\leq 1/\lambda$ is strong enough to exclude the unconstrained minimum $x_{\text{unc}}^*$ of $f(x)$. This may improve the generalization and stability of the solution while sacrificing the training loss. 

2) If $\lambda$ is sufficiently small to include the unconstrained minimum $x_{\text{unc}}^*$ in the constrained set, 
it does not alter the final solution. 
In this case, the main role of weight decay is to speed up the convergence because Phase 1 brings the solution into the constrained set with a linear rate. Hence, the ideal choice of $\lambda$ is $\lambda = 1/\norm{x_{\text{unc}}^*}_\infty. $ 

In Figure~\ref{fig:converged_loss} we plot \lion{}'s performance with different $\lambda$. The right plot confirms that larger $\lambda$ results in faster convergence but might sacrifice the performance. The left plot shows that there exists an optimal $\lambda$ (=0.56), beyond which the training loss starts to increase.

\paragraph{Going Beyond Lion} 
Different $\phi$ 
yield optimization with different convex constraints and/or regularizations. 
For example, 
using the  $\ell_p$ norm  $\phi(x) = \norm{x}_p$ 
yields a constraint on the dual norm $\norm{x}_q\leq 1/\lambda$ where $1/p + 1/q = 1$ (Table~\ref{tab:phiexamples}, Line~\mycircle{2});  
zeroing out the coordinates 
with small magnitude corresponds to introducing an $\ell_1$ regularization (Line~\mycircle{3}) or $\ell_1$ constraint (\mycircle{4}),  
which is useful for sparse learning; 
replacing $\dd\phi(x) = \sign(x)$  with a continuous function
would introduce an extra regularization term on the loss (e.g., \mycircle{5}). 
This work will focus on building the basic theoretical framework, 
and leave the vast opportunities of practical applications as future directions. 
}

\renewcommand{\arraystretch}{1.5} 
\setlength{\tabcolsep}{10pt} 
\begin{table} 
\centering 
\scalebox{.9}{\small 
\begin{tabular}{c|c|c|c} 
\hline 
Line ID & $\phi(x)$ & $\dd \phi(x)$ & $\min_x f(x) + \phi^*(x)$ \\ \hline 
\mycircle{1}&$\norm{x}_1$ &  $\sign(x)$ &  
$\min f(x) ~~s.t.~~\norm{x}_\infty \leq 1$
\\\hline 
\mycircle{2} &$\norm{x}_p$ & $\frac{\sign(x) \abs{x}^{p-1}}{\norm{x}_p^{p-1}}$ & 
$\min f(x) ~~s.t.~~\norm{x}_q  \leq 1 $
\\ \hline 
\mycircle{3}
&$\sum_i \max(\abs{x_i}-e, 0) $ & 
$\sign(x) \ind(\abs{x} > e) $ & 
$\min f(x) + e \norm{x}_1 ~~s.t.~~  \norm{x}_\infty\leq 1 $ \\ \hline
\mycircle{4}
&$\sum_{i\leq i^{cut}} \abs{x_{(i)}}$ 
& $\sign(x) \ind(\abs{x}>\abs{x_{(i^{cut})}}) $
& 
$\min f(x)  ~~s.t.~~  \norm{x}_1\leq i^{cut},~~\norm{x}_\infty\leq 1 $ 
\\ \hline 
\mycircle{5}
&$\sum_i \mathrm{huber}_e(x_i)$ 
& $\mathrm{clip}(x, -e, e) / e $ 
& $\min f(x)+\frac{e}{2}  \norm{x}_2^2~~s.t.~~ \norm{x}_\infty < 1$ \\ \hline 
 \end{tabular}
 }
\caption{Examples of $\phi$ and  $\dd \phi$, and the optimization problems they solved (we set $\gamma = \lambda = 1$ for simplicity). 
We assume $x  = [x_1,\ldots, x_d]\in\RR^d$ and 
$\abs{x_{(1)}} \geq \abs{x_{(2)}} \geq \cdots $ is a monotonic sorting of the elements of $x$, and $i^{cut}$ is an integer in $\{1,\ldots, d\}$. 
The Huber loss is $\mathrm{huber}_e(x_i) = \ind(\abs{x_i}\geq e) (\abs{x_i} - \frac{e}{2}) 
+ \ind(\abs{x_i} < e) \frac{1}{2e} x_i^2$, $e > 0$.  
See Appendix~\ref{sec:Kexamples} for more examples. 
} 
\label{tab:phiexamples}
\end{table}

\paragraph{Outline}
The rest of the paper is organized as follows. 
Section~\ref{sec:convex} introduces 
preliminaries on convex functions. 
Section~\ref{sec:continuous} analyzes the continuous-time Lion-$\phi$ dynamics and discusses connections with existing algorithms. 
Section~\ref{sec:discrete} presents the discrete-time analysis. \rebuttal{Section~\ref{sec:experiment} presents experiments that study and verify the behavior of using different $\phi$s.}

\section{Preliminaries on Convex Functions}
\label{sec:convex}

\label{sec:convexity}
Assume $\phi \colon \RR^d\to \RR$ is convex. A vector $u\in \RR^d$ is said to be a subgradient of $\phi$ at $x$, denoted as $u \in \partial \phi(x)$, if 
$$
\phi(y) - \phi(x) \geq  u\tt (y-x), ~~~~\forall y \in \RR^d. 
$$
With an abuse of notation, we use $\dd \phi(x)$ to denote a subgradients of $\phi$, that is, $\dd\phi(x) \in \partial \phi(x)$. 
When $\phi$ is differentiable at $x$, there is an unique subgradient $\dd \phi(x)$
which coincides with the  regular derivative. 

The  conjugate function $\phi^*$ of $\phi$ is defined as 
$$
\phi^*(x) = \sup_{z\in\RR^d} (x \tt z - \phi(z)). 
$$
Hence, by definition, we have the following Fenchel-Young inequality:
\bbb \label{equ:youngineq} 
\phi(x) + \phi^*(y) \geq x \tt y,~~~\forall x, y. 
\eee 
The conjugate function $\phi^*$ can take values in the extended real set $\bar\RR = \RR\cup\{\pm\infty\}$, and $\phi^*$ is always closed and convex, even when $\phi$ is not. Recall that a function $f$ is said to be closed if for each $b\in \RR$, its sublevel sets $\{x \colon f(x) \leq b\}$ is a closed set. 

If $\phi$  is closed and convex, we have $\phi^{**} = \phi$,
and 
\bbb \label{equ:inversemap} 
y \in  \partial \phi(x) 
&& \iff   &&
x \in  \partial \phi^*(y)
&& \iff &&
\phi(x) + \phi^*(y) = x \tt y. 
\eee  
When $\phi$ and $\phi^*$ are differentiable, 
\eqref{equ:inversemap} suggests that $\dd \phi$ and $\dd\phi^*$ is a pair of inverse maps: $\dd\phi(\dd\phi^*(x)) = x$. 
Combining \eqref{equ:youngineq}
 and \eqref{equ:inversemap}, we get 
 $\min_{m} \phi(m) + \phi^*(x) - x \tt m = 0$, 
 which yields $F(x) = \min_{m} H(x, m)$. 
We refer to \citet{rockafellar1997convex} 
for a systematic introduction to convex functions. 

A key property of any subgradient 
$\dd\phi$ and $\dd\phi^*$ is that they are monotonic maps, which plays a crucial rule in our results. 

\begin{lem}\label{lem:key00}
Assume $\phi, \phi^*$ is a closed convex conjugate pair and $\dd\phi$, $\dd\phi^*$ are their subgradients, we have 
\bbb \label{equ:keyineqs} 
(\dd\phi(x) - \dd\phi(y))\tt (x-y) \geq 0, 
&&
(\dd\phi(x) - y) \tt 
(x - \dd\phi^*(y)) \geq 0. 
\eee  
\end{lem}
See Appendix~\ref{sec:convexity} for the proof. 
These two inequalities are crucial because they allow us to identify vectors that have a non-negative inner product with a given direction to achieve monotonic descent in optimization.

\begin{exa}
In the case of Lion, we take $\phi(x) = \norm{x}_1$ with  
$\dd\phi(x) = \sign(x) $, and 
\bb 
%
\phi^*(y) = \begin{cases}
0 & \text{if $\norm{y}_\infty \leq 1$} \\
+\infty & \text{if $\norm{y}_\infty > 1$} 
\end{cases}, 
&&
[\dd\phi^*(y)]_i= \begin{cases}
0 & \text{if $\abs{y_i} \leq 1$} \\
+\infty & \text{$y_i> 1$} \\
-\infty & \text{$y_i <-1$}.  
\end{cases} 
\ee 
One can verify that the inequalities in \eqref{equ:keyineqs} hold (even though the values on the left side can be $+\infty$).
The Lyapunov function in \eqref{equ:H1} becomes
\bb 
H(x,m) =
\begin{cases} f(x) + \frac{1-\varepsilon \gamma}{1+\varepsilon \lambda }(\norm{m}_1 -\lambda x \tt m) & \text{if $\norm{x}_\infty \leq 1$}  \\
+\infty & \text{if $\norm{x}_\infty >1$}. 
\end{cases}
\ee 
\end{exa} 

\section{Main Result: Continuous-Time} 
\label{sec:continuous} 

We study the continuous-time 
Lion-$\phi$ dynamics \eqref{equ:lionode}, and discuss its connection to existing algorithms listed in Table \ref{tab:lionk}.   
We defer the detailed proofs to Appendix~\ref{sec:proofsmain}, 
but outline 
a novel \emph{implicit Hamiltonian + descent decomposition} 
that underpins the construction of the Lyapunov function $H(x,m)$. 

\begin{thm}\label{thm:main}
Let $(x_t, m_t)$ be a continuously differentiable trajectory of the Lion-$\phi$ ODE \eqref{equ:lionode}, 
where $\phi$ is differentiable convex with conjugate $\phi^*$.  
Assume
$\alpha,\gamma,\lambda,\varepsilon > 0$ and $\varepsilon \gamma \leq 1$. 

1) \textbf{[Phase 1]} 
Define $\dist(\lambda x_t, \dom \phi^*) =  \inf_{z \in \dom \phi^*}\norm{z - \lambda x_t}$ w.r.t. any norm $\norm{\cdot}$. We have 
$$
\dist(\lambda x_t, \dom \phi^*) \leq \exp(\lambda (s-t)) ~\dist(\lambda x_s, \dom \phi^*),~~~~\forall  0 \leq s \leq t. 
$$ 
Hence, $\lambda x_t$ converges linearly to set $\dom \phi^*$ and stays within $\dom \phi^*$ once it enters it.

2) \textbf{[Phase 2]}  
When $H(x,m)$  in \eqref{equ:H1} is finite and continuously differentiable, it is decreased monotonically along the trajectory: 
$$
- \ddt H(x_t, m_t) =  \Delta (x_t, m_t)
\defeq   \frac{\lambda +\gamma}{1+\varepsilon \lambda }
\Delta_1(x_t, \tilde m_t)
+  \frac{1-\varepsilon\gamma}{1+\varepsilon \lambda  }
\Delta_2(m_t, \tilde m_t)
 \geq 0,
$$
where we define $\tilde m_{t} = m_t - \varepsilon (\alpha \dd f(x_t) + \gamma m_t)$, and 
\bbb \label{equ:ineqxm} 
\begin{split}
& \Delta_1(x_t, \tilde m_t) = (\tilde m_t- \dd\phi^*(\lambda x_t))\tt 
(\dd\phi(\tilde m_t) - \lambda x_t)\geq0, \\ 
& \Delta_2(m_t, \tilde m_t) = \frac{1}{\varepsilon} (\tilde m_t - m_t)  \tt (\dd\phi(\tilde m_t) - \dd\phi(m_t))\geq 0. 
\end{split} 
\eee  
3) \textbf{[Stationarity]}
Assume $\dd\phi^*$ is strictly monotonic. 
All the accumulation points of $(x_t, m_t)$ as $t\to+\infty$ are stationary points of the objective function $F(x) = \alpha f(x) + \frac{\gamma}{\lambda }\phi^*(\lambda x),$ and satisfy $\lambda x \in \dom \phi^*$. 
\end{thm} 

$\Delta(x_t, m_t)$ can be viewed as an indication of 
the stationarity of the system. If $H(x_0,m_0)$ is finite and  $H_b \defeq \inf_{x,m}H(x,m) > -\infty$,
we have $\frac{1}{T}\int_0^T \Delta(x_t, m_t)\d t \leq \frac{H(x_0,m_0)-H_b}{T} \to 0$ when $T\to +\infty$. 

\begin{proof}[Proof Sketch]
See Appendix~\ref{sec:proofsmain} for the full proof. 
The original discovery of the Lyapunov function 
was made possible by 
starting from the inequalities in \eqref{equ:ineqxm} 
as guaranteed by Lemma~\ref{lem:key00}, and working backwards with some guesswork. 
The following is a simplified proof that highlights the essential mathematical structure that makes $H(x, m)$  Lyapunov.  
Define   
\bb 
 \dot x = V_x(x,m) \defeq \dd \phi(\tilde m) - \lambda x , &&
 \dot m = V_m(x,m) \defeq  - \alpha\dd f(x) - \gamma m = \frac{\tilde m - m}{\varepsilon}
\ee  
and related 
\bb 
 \hat V_x(x,m) = \tilde m - \dd \phi^*(\lambda x) , 
&&  \hat V_m (x,m) =  \dd\phi(\tilde m) - \dd \phi(m).
\ee 
The $\hat V_x$ and $\hat V_m$ have two critical properties: 

1) By Lemma~\ref{lem:key00}, $\hat V_x$ and $\hat V_m$ have non-negative inner products with $V_x, V_m$, respectively: 
\bb 
\hat V_x(x,m) \tt  V_x(x,m) \geq 0,  && 
\hat V_m(x,m) \tt V_m(x,m) \geq 0,~~~~~~~\forall x, m. 
\ee 
2) By 
Lemma~\ref{lem:decompH} in Appendix~\ref{sec:proofsmain}, 
the gradients of $H$ can be decomposed as follows: 
\begin{align} 
\begin{split}
\dd_x H(x, m) = \blue{ - \eta'\hat V_x} \med{- \eta V_m }\\ 
\dd_m H(x, m) =  \blue{- \eta \hat V_m}  \med{+ \eta V_x },
\end{split} && 
\textbf{(Implicit Hamiltonian + Descent)}
\label{equ:HV}
\end{align}  
where $\eta = \frac{1-\varepsilon\gamma}{1+\varepsilon\lambda}$ 
and $\eta' = \frac{\gamma+\lambda }{1+\varepsilon\lambda}$. 
We call \eqref{equ:HV} an 
\emph{``implicit" Hamiltonian + descent} decomposition, in connection with the Hamiltonian + descent decomposition we introduce in sequel. 

Then we have, 
\bb 
\ddt H(x_t, m_t)
& = \dd_x H\tt V_x 
+ \dd _m H \tt V_m   = (\blue{-\eta' \hat V_x} \med{ - \eta V_m})\tt V_x 
+ (\blue{- \eta \hat V_m} \med{+ \eta V_x})\tt V_m  \\
&  = - (\blue{\eta' \hat V_x\tt V_x + \eta \hat V_m \tt V_m}) \leq 0. 
\ee 
The key here is that the cross term $\med{\eta V_x\tt V_m}$ is canceled, leaving only the negative terms. 
The convergence property uses LaSalle's invariance principle; see Appendix~\ref{sec:proofsmain} for details.   
\end{proof}

\paragraph{Hamiltonian + Descent Decomposition}
The decomposition structure \eqref{equ:HV} is a key characterization of Lion-$\phi$ ODE. An interesting remark is that $H(x,m)$ is also Lyapunov if we have the following 
\emph{Hamiltonian + descent} structure  
\citep{maddison2018hamiltonian, o2019hamiltonian} 
in which 
the roles of $[\dd_xH, \dd_m H]$ and $[V_x, V_m]$ 
in \eqref{equ:HV} are switched:  
\bbb \label{equ:HVdual}
\begin{split}
& V_x = \blue{- \hat H_x} \med{- \eta \dd_m  H} \\ 
& V_m  =  \blue{ - \hat H_m} \med{+ \eta \dd_x H}, 
\end{split}
&& 
\textbf{(Hamiltonian + Descent)}
\eee  
where $\hat H_x, \hat H_m$ are two vector fields satisfying 
 $\hat H_x  \tt (\dd _x H) \geq 0$ and $\hat H_m \tt (\dd _m H) \geq 0$, then  
\bb 
\ddt H(x_t, m_t)
& = \dd_x H\tt V_x 
+ \dd _m H \tt V_m   = \dd_x H \tt (\blue{-\hat H_x} \med{ - \eta \dd_m H}) 
+ \dd_m H \tt (\blue{-  \hat H_m} \med{+ \eta H_x})\\
&  = - (\blue{ \hat H_x\tt (\dd_x H) + \hat H_m \tt(\dd_m H)}) \leq 0. 
\ee 
The structure in \eqref{equ:HVdual} can be intuitively viewed as a generalized damped Hamiltonian system with $H(x,m)$ as the total energy, 
where $[-\hat H_x, - \hat H_m]$ serves a damping force that monotonically  decreases the total energy, and 
$[-\dd_m H, \dd_x H]$ is the Hamiltonian vector field 
which preserves the energy but introduces an inertia-like effect into system. 
One can easily verify \eqref{equ:HVdual} on the classical Polayk's momentum. 
The more general idea is explored in the Hamiltonian descent method of \citep{maddison2018hamiltonian, o2019hamiltonian},  
which considers systems of structure \eqref{equ:HVdual} for the separatiable Hamiltonian of form $H(x,m) = f(x) + \phi(m)$ with 
$\hat H_x = 0$. 
%
%
In contrast, \eqref{equ:HV} do not seem to have a  clear physical interpretation, yet provides a handy tool for understanding the general Lion-$\phi$ dynamics. 
Some special cases of Lion-$\phi$, 
such as when $\lambda =0$ or $\varepsilon = 0$, 
can also be alternatively viewed from the Hamiltonian + descent structure as shown in Section~\ref{sec:existing}.  


\subsection{Connection with Existing Algorithms} 
\label{sec:existing}

What makes Lion-$\phi$ unique is the combination of 
the gradient enhancement ($\varepsilon > 0$), 
the decoupled weight decay ($\lambda > 0$), 
and the momentum damping ($\gamma > 0$), 
the use of reshaper function $\dd \phi(\cdot )$.  
We discuss the effects of these elements in connection to existing algorithms as shown in Table~\ref{tab:lionk}. 

\paragraph{Lion-$\phi$ Without Weight Decay} 
When $\lambda = 0$ and $ \dd \phi^*(0)=0$, 
we have $\lim_{\lambda \to 0}\frac{1}{\lambda } \phi^*(\lambda x) = \dd \phi(0)\tt x=0$, and 
the Lyapunov function can be defined as 
\bb 
H(x, m) = \alpha f(x) + (1-\varepsilon \gamma) \phi(m), 
\ee 
for which we have 
\bb 
- \ddt H(x_t, m_t) = 
   \gamma 
   \dd\phi(\tilde m_t) \tilde m_t  
+ \frac{(1-\varepsilon \gamma)}{\varepsilon} 
(\tilde m_t - m_t )\tt (\dd\phi(\tilde m_t) - \dd\phi(m_t)) \geq 0.
\ee 
In this case, the algorithm solves $\min_x  f(x)$, without the regularization term $\phi^*(\lambda x)$. 

Interestingly, in this case ($\lambda =0$) and $1-\varepsilon \gamma >0$, 
there exists a second Lyapunov function: 
\bbb 
\label{equ:hsecond}
\tilde H(x, m) = 
\alpha f(x)  + \frac{1}{1-\varepsilon\gamma}\phi((1-\varepsilon\gamma) m ), 
\eee  
with which the Lion-$\phi$ ODE ($\lambda=0$) can be  decomposed in the form of  \eqref{equ:HVdual}, as a sum of a Hamiltonian vector field and a descent direction: 
\bb
\begin{bmatrix}
\dot x_t \\
\dot m_t 
\end{bmatrix}  
= \blue{\underbrace{\begin{bmatrix}
+ \dd_m \tilde  H(x_t, m_t) \\
- \dd_x \tilde H(x_t, m_t)
\end{bmatrix}}_{\text{Hamiltonian}}}    
- \med{\underbrace{\begin{bmatrix}
\dd  \phi(\tilde m_t^0) - \dd  \phi(\tilde  m_t) \\
\gamma m_t
\end{bmatrix}}_{\text{Descent}}},
\ee 
where $\tilde m_t^0 = (1-\varepsilon \gamma) m_t$ 
and hence $\tilde m_t^0 - \tilde m_t = \varepsilon \alpha \dd f(x_t)$. 
If $m=0$ is a minimum of $\phi(m),$ one can show that the second component above is a descent direction of $\tilde H(x,m)$ in \eqref{equ:hsecond}, with 
\bb 
- \ddt \tilde H(x_t, m_t) 
 =   \gamma 
\dd  \phi(\tilde m_t^0) \tt m_t  +  \frac{1}{\varepsilon}(\tilde m_{t}^0 - \tilde m_t) \tt  (\dd \phi(\tilde m_t^0) - \dd  \phi(\tilde m_t))  \geq 0,
\ee 
See Appendix~\ref{sec:lionsolved} for details.


\paragraph{Lion-$\phi$ Without Momentum Damping}
When $\gamma =0$, we have  
$$
H(x, m) = \alpha f(x) + \frac{1}{1+\varepsilon \lambda } 
(\phi^*(x) + \phi(m) - \lambda x \tt m ), 
$$
Because $\min_m (\phi^*(x) + \phi(m) - \lambda x \tt m ) = 0$, 
the algorithm also corresponds to solving $\min_x  f(x)$ without 
regularization $\phi^*(\lambda x)$. 

It is interesting to see that the weight decay and momentum damping 
play a somewhat symmetric role, because 
 turning off either one of it turns off the regularization term $\phi^*(\lambda x)$. 
In particular, if $\phi(x) = \norm{x}_2^2/2$, the Lion-$\phi$ 
ODE can be rewritten into a second-order ODE: 
\bbb \label{equ:odephi2} 
\ddot x_t + (\lambda + \gamma) \dot x_t + \varepsilon \alpha \dd^2f(x_t) \dot x_t + 
  \gamma \lambda x_t + 
\alpha \dd f(x_t)  = 0, 
\eee 
in which the role of $\gamma,\lambda$ are symmetric.
Equation \eqref{equ:odephi2} coincides the high-resolution ODE 
in \cite{shi2021understanding} for minimizing $F(x) = \alpha f(x) + \gamma\lambda \norm{x}^2_2/2$, 
which is a high resolution continuous time limit of Nesterov momentum. 
The hessian-based damping term $\dd^2 f(x_t) \dot x_t$ plays 
a key role for acceleration phenomenon \citep[see e.g.,][]{shi2021understanding, attouch2016fast}.   
When we turn off the gradient enhancement ($\varepsilon = 0$), 
then we get ODE for Ployak momentum. 

Interestingly, if we set $\lambda = \gamma = 0$, but $\varepsilon >0$, 
 ODE \eqref{equ:odephi2} still serve to minimize $f(x)$, due to the Hessian damping term.  
 %
  
\paragraph{Lion-$\phi$ without Gradient Enhancement}   
When $\varepsilon =0$, we have 
\bb
H(x, m) = \alpha   f(x)   + 
\frac{ \gamma }{\lambda}
\phi^*(\lambda x) + 
(\phi^*(\lambda x) +  \phi(m)   - \lambda 
m\tt  x ), 
\ee
and $\Delta_2(m, \tilde m) = 0$, 
$$
\Delta(x, m) 
 = ({\lambda +\gamma}) \Delta_1(x, m) = 
 ({\lambda +\gamma})  ( m - \dd\phi^*(\lambda x))\tt 
(\dd\phi(m) - \lambda x). 
$$
In this case, minimizing $H(x, m)$ still yields the minimization of $F(x)$.  
Hence, the choice of $\varepsilon$ does not alter the objective function. 

Moreover, with $\varepsilon = 0$, one can conveniently decompose the velocity field in the form of  \eqref{equ:HVdual}, as a sum of a Hamiltonian vector field and  mirror descent direction: 
\bb 
\begin{bmatrix}
\dot x_t \\
\dot m_t
\end{bmatrix} 
& = 
\blue{\underbrace{\begin{bmatrix} \blue{+ \dd_m H(x_t, m_t)}\\ 
 \blue{-\dd_x H(x_t, m_t )} 
 \end{bmatrix} }_{\blue{\text{
 Hamiltonian}}}} 
 -  
 \med{\underbrace{\begin{bmatrix} 0 \\
 (\gamma + \lambda ) {(m_t - \dd\phi^*(\lambda x_t))} 
 \end{bmatrix}}_{\text{
  Descent}}}.
\ee 
This system can be shown  to be equivalent to the Hamiltonian descent 
system for composite objects of \cite{o2019hamiltonian}. 
Further, if $\lambda = 0$, it reduces to 
 the conformal Hamiltonian system  \citep[e.g.,][]{maddison2018hamiltonian, mclachlan2001conformal}.





\paragraph{Mirror Descent and Frank-Wolfe} 
If  $\varepsilon \gamma =1 $, Lion-$\phi$ reduces to 
$$
\dot x_t = \dd\phi(-\varepsilon \alpha \dd f(x_t)) - \lambda x_t, 
$$
which can be shown to  be equivalent to
the Frank-Wolfe algorithm for minimizing $F(x) = \alpha f(x) + \frac{\gamma}{\lambda} \phi^*(\lambda x)$.

When $\varepsilon \gamma = 1$, and $\lambda = 0$ with $\dd\phi(x) =0$ iff $x=0$, 
Lion-$\phi$ reduces to $\dot x_t = \dd\phi(- \varepsilon \alpha \dd f(x_t))$, which  
is dual space conditioning \citep{maddison2021dual},  or a variant of 
mirror descent for $\min_x f(x)$.  See Appendix~\ref{sec:fw} for more discussion. 


\paragraph{Accelerated Mirror Descent}  
The accelerated mirror descent of \citet{krichene2015accelerated}   is 
\bb 
\dot x_t = \lambda_t (\dd\phi(m_t) - x_t), && 
\dot m_t = -\alpha_t \dd f(x_t),  
\ee 
which is shown to exhibit an acceleration behavior 
for minimizing a convex $f$ (without the $\phi^*$ regularization)
when $\alpha_t = t/r$ and $\lambda_t = r/t$ 
and $r \geq 2$. This 
can be viewed as Lion-$\phi$ ODE with $\gamma=0,\varepsilon=0$ and but a special time-dependent coefficient. 

\section{Discrete Time Analysis} \label{sec:discrete}

We now present a result on the discrete-time Lion-$\phi$ 
parallel to the continous-time results in 
Theorem~\ref{thm:main}, but work for non-differentiable convex functions  $\phi$. 
We analyze a slight reform of \eqref{equ:lionK}: 
\bbb \label{equ:finiteupdate0} 
\begin{split}
m_{t+1} & = \btwo m_t  - (1-\btwo) \dd f(x_t) \\ 
\tilde m_{t+1} & = \bone m_{t} - (1-\bone) \dd f(x_t) \\ 
x_{t+1} & = x_t + \lr
(\dd \phi(\tilde m_{t+1}) - \lambda x_{t+1}), 
\end{split} 
\eee  
in which we use an implicit scheme for the $x_t$-update, replacing $\lambda x_t$ with $\lambda x_{t+1}.$ It is equivalent to the explicit scheme in \eqref{equ:lionK} with $\epsilon $ replaced by $\epsilon' =  \frac{\lr}{1+\lr \lambda }$. 
\begin{thm}\label{thm:discretemain}
Assume 
$f\colon \RR^d\to \RR$ is $L$-smooth, and $\phi\colon \RR^d\to \RR$ is closed and convex, and $\dd\phi$ is a subgradient of $\phi$.  
Assume $\bone,\btwo \in (0,1)$, and $\btwo>\bone$, and $\epsilon, \lambda > 0$. 

1) For any two non-negative integers $s \leq t,$ we have 
$$
\dist(\lambda x_t, \dom \phi^*) \leq \left (\frac{1}{1+\epsilon \lambda }\right)^{s-t} \dist(\lambda x_s, \dom \phi^*), ~~~\forall s\leq t.
$$

2) Define the following Lyapunov function: 
$$
H(x, m ) = f(x) + \frac{1}{\lambda } \phi^*(\lambda x) + \frac{\bone}{\epsilon \lambda  (1-\bone) + (1-\btwo)}(\phi^*(\lambda x) + \phi(m) - \lambda x \tt m ), 
$$
and 
\bb 
\Delta_t^1  & =(\dd \phi(\tm_{t+1}) - \lambda x_{t+1}) 
\tt 
( \tm_{t+1}  - \dd\phi^*(\lambda x_{t+1})) \geq 0, \\ 
\Delta^2_t & = 
 (\dd\phi(\tilde m_{t+1}) - \dd\phi(m_{t+1}))\tt  ( \tm_{t+1} - m_{t+1}) \geq 0, 
\ee 
where $\dd\phi^*$ is a subgradient of $\phi^*$. 
Then we have 
\bb 
H(x_{t+1}, m_{t+1}) - H(x_t, m_t)
\leq 
- \epsilon \Delta_t  + \frac{L\epsilon^2}{2} \norm{\dd \phi(\tm_{t+1}) - \lambda x_{t+1}}_2^2, 
\ee 
where $\Delta_t = a  \Delta_t^1 
+ b \Delta_t^2$, with 
\bb 
a = \frac{\bone  }{ \lr \lambda (1-\bone) + (1-\btwo) } + 1 \geq 0, &&  
b = \frac{\bone(1-\btwo)}{\lr \lambda (\btwo-\bone)(\lr \lambda (1-\bone) + (1-\btwo)) }\geq 0. 
\ee 
Hence, a telescoping sum yields 
$$
\frac{1}{T}\sum_{t=0}^{T-1} \Delta_t 
\leq \frac{H(x_0,m_0) - H(x_{T}, m_T)}{\epsilon T} + \frac{L\epsilon }{2} B_T, 
$$
where $B_T = \frac{1}{T}\sum_{t=0}^{T-1} \norm{\dd\phi(\tm_{t+1}) - \lambda x_{t+1}}^2_2$. 
\end{thm} 
\rebuttal{
The result above shows that 
$\frac{1}{T}\sum_{t=0}^{T-1} \Delta_t $ decays with an $O(\frac{1}{\epsilon T} + \epsilon)$ rate, if $B_T$ is a finite upper bound.  
This reduces to the continuous-time result of $\frac{1}{t}\int_0^t \Delta(x_s,m_s)\d s = O\left (\frac{1}{t}\right )$ when the step size $\epsilon$ converges to zero. 

If $\mathcal K$ is smooth, it is possible to improve the discrete-time rate to $O\left (\frac{1}{\epsilon T}\right )$ with standard arguments based on the proof of Theorem~\ref{thm:discretemain}.  
Hence, the impact of the non-differentiability of $\phi$ contributes to the $O(\epsilon)$ term, which suggests that the algorithm converges upto an $\epsilon$ accuracy. 
This is an typical phenomenon in optimization with non-smooth objectives (like sub-gradient descent) or non-smooth update (like signed GD). 
Because in practice the step size is  small or decaying, the $O(\epsilon)$ term may not have a substantial impact for practical performance. 
}



\rebuttal{
\section{Experiments on Different $\phi$}
\label{sec:experiment}
This section provides a preliminary investigation on the behaviors of \lion{}-$\phi$ with different $\phi$. We experiment with the $\phi$s listed in Table~\ref{tab:phiexamples} on the toy example shown in Figure~\ref{fig:traj-1} to confirm the behavior follows exactly as what the theory predicts. Then we focus on the \lion{}-$\ell_p$ optimizer with general $p \in [1, 2]$ since it is the most straightforward extension of the original \lion{} (with $p=1$).

\subsection{\lion{}-$\phi$s on the Toy Example}
In the following, we plot the behavior of different \lion{}-$\phi$s on the toy example shown in Figure~\ref{fig:traj-1}. For each $\phi$, we draw the optimization trajectory using the corresponding optimizer, the loss $f(x)$, and the corresponding constraint (e.g., the norm of $x$) v.s. iteration. The results are shown in Figure~\ref{fig:different_k}.
\begin{figure}[t!]
    \centering
    \includegraphics[width=\textwidth]{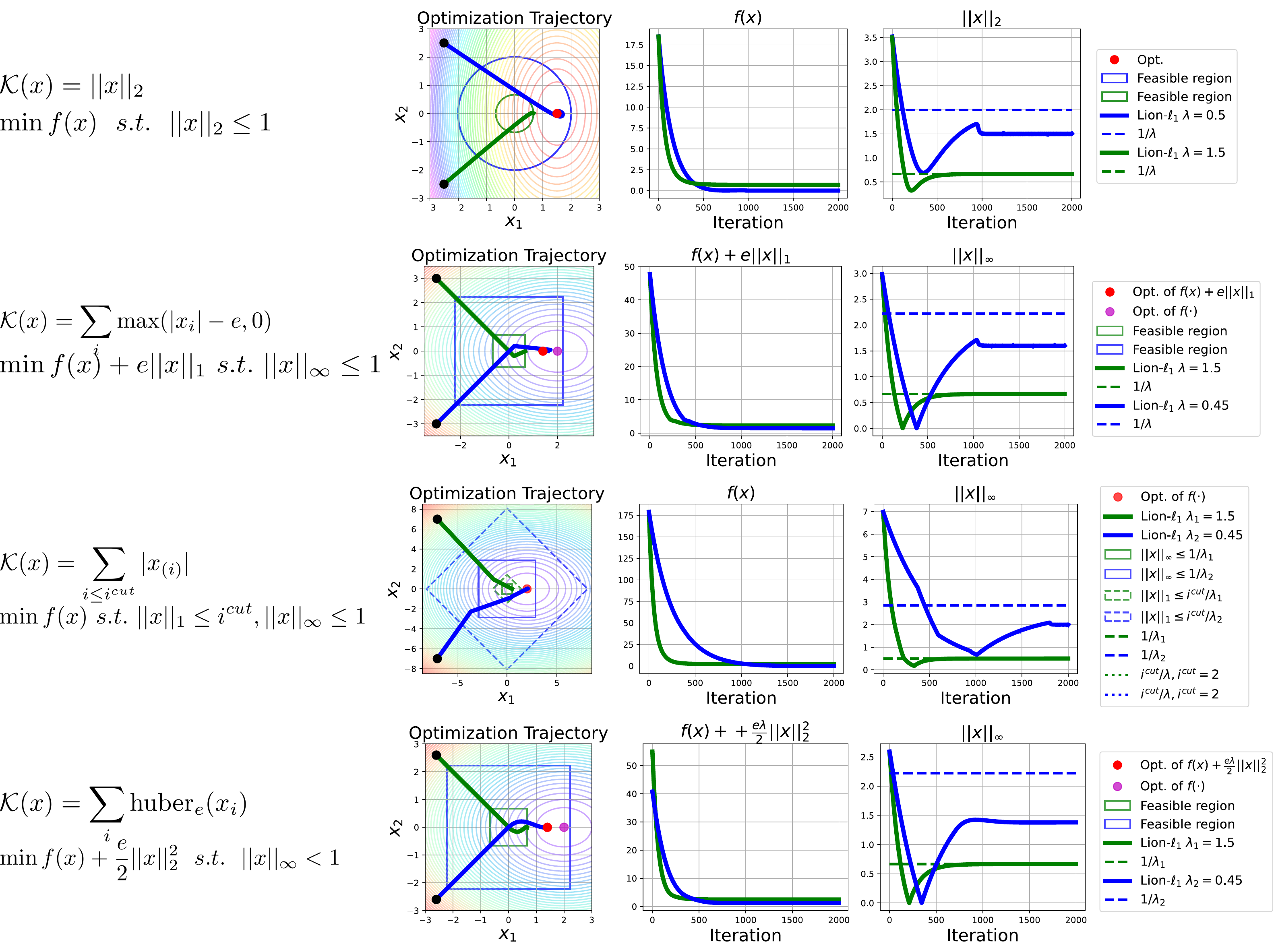}
    \vspace{-5pt}
    \caption{The behavior of \lion{}-$\phi$ with different $\phi$s from Table~\ref{tab:phiexamples}. The \textcolor{blue}{blue} trajectory always reaches the optimum as the optimum is included in the constraint. The \textcolor{teal}{green} trajectory converges to the boundary of the constraint.}
    \label{fig:different_k}
    \vspace{-5pt}
\end{figure}
\paragraph{Observation} From Figure~\ref{fig:different_k}, one can observe that for $\phi(x)= \norm{x}_2$, the constraint is a circle. For $\phi(x) = \sum_i \max(\abs{x_i} - e, 0)$, 
an additional $\ell_1$ regularization is introduced in addition to the 
$\ell_\infty$ constraint, which encourages sparse solutions. 
When $\phi(x) = \sum_{i \leq i^{cut}} |x_{(i)}|$, it enforces an $\ell_1$ constraint (rather than regularization) in addition to the $\ell_\infty$ constraint. 
The $\phi(x) = \sum_i \text{huber}_e(x_i)$ 
 introduces an $\ell_2$ regularization effect in addition to 
$\ell_\infty$ constraint. All optimization trajectories closely match what the theory predicts.

\subsection{Lion-$\ell_p$ for ImageNet and Language Modeling}
Lion-$\ell_p$ corresponds to $\phi(x) = \norm{x}_{p}$, $p\geq 1$ and amounts to solving $\min_x f(x)~ s.t. ~\norm{x}_q \leq 1/\lambda$ where $1/p + 1/q =1$. In Figure~\ref{fig:imagenet-constraint}, we plot how the  parameter norms (e.g., $||\cdot||_\infty$ when $p=1$ and $||\cdot||_2$ when $p=2$) change over training iterations. In Figure~\ref{fig:imagenet}, we compare the performance of using \lion{}-$\ell_p$ with different $p$, on ImageNet~\citep{ILSVRC15} and Language Modeling tasks, using ResNet-50, Vision Transformer (ViT)~\citep{dosovitskiy2020image}, and the GPT-2 model~\citep{radford2019language}. 

\paragraph{Experiment Setting} For the ImageNet training, we follow the standard PyTorch ImageNet training code.\footnote{\url{https://github.com/pytorch/examples/blob/main/imagenet/main.py}.} We train the ResNet-50 and the ViT-B/16 model using batch size 1024 and cosine learning rate scheduler. For GPT-2 training, we follow the HuggingFace code\footnote{\url{https://huggingface.co/gpt2}}, train it on OpenWebText\footnote{\url{https://huggingface.co/datasets/Skylion007/openwebtext}} using cosine learning rate scheduler.

\paragraph{Observation} 
From Figure~\ref{fig:imagenet-constraint}, we observe that even on deep neural networks like ViT~\citep{dosovitskiy2020image}, ResNet~\citep{he2016deep}, and GPT-2~\citep{radford2019language}, the behavior of the \lion{}-$\phi$ optimizers strictly follow what the theory predicts. From Figure~\ref{fig:imagenet}, we observe that \lion{}-$\ell_1$ (the original \lion{} optimizer) performs better than \lion{} with other $p$ on ImageNet when ViT is used, and on language modeling with the GPT-2 model. The plot indicates a trend that smaller $p \in [0, 1]$ results in better training efficiency. However, the trend is reversed when ResNet-50~\citep{he2016deep} is used on ImageNet. Therefore, this indicates that the choice of $\phi$ might depend on the underlying neural architecture. Based on the empirical observation, we conjecture that \lion{}-$\ell_1$ performs well among all \lion{}-$\ell_p$ on the transformer architecture, which is consistent with the fact that \lion{}-$\ell_1$ is found by an evolutionary search using the transformer architecture~\citep{chen_symbolic_2023}.

\begin{figure}[h!]
    \centering
    \includegraphics[width=\textwidth]{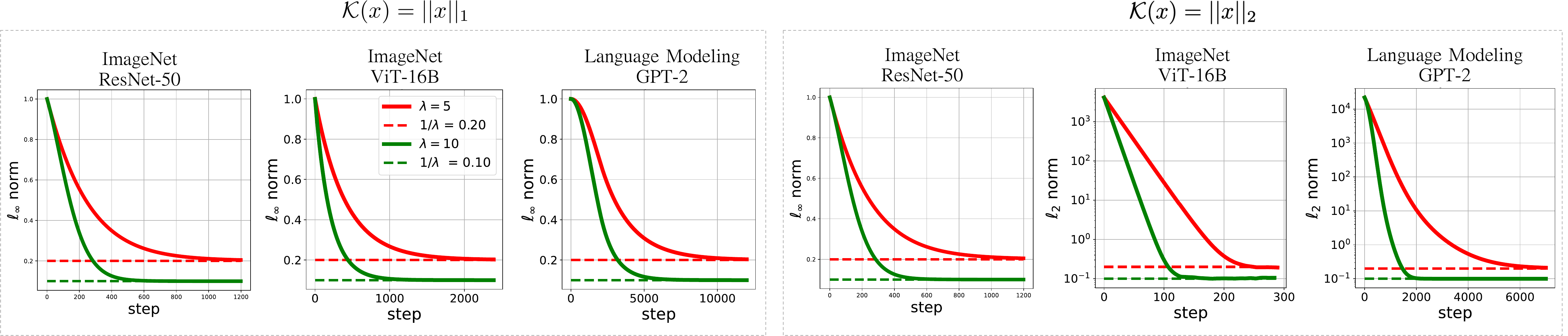}
    \vspace{-10pt}
    \caption{Constraint verification for \lion{}-$\ell_1$ and \lion{}-$\ell_2$ on ImageNet and Language Modeling tasks, using the ResNet-50, ViT-B/16 and the GPT-2 architectures.}
    \label{fig:imagenet-constraint}
    \vspace{-10pt}
\end{figure}
\begin{figure}[h!]
    \centering
    \includegraphics[width=\textwidth]{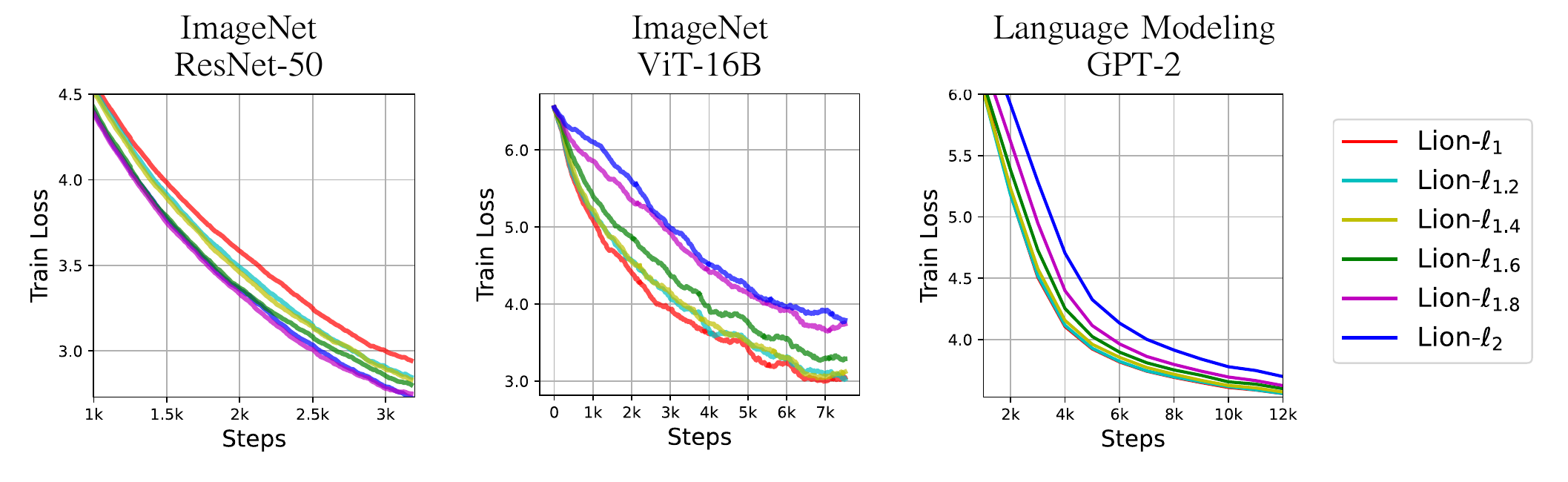}
    \vspace{-20pt}
    \caption{Performance of \lion{}-$\ell_p$ with different $p$, on ImageNet~\citep{ILSVRC15} (left 2 figures) and Language Modeling (right), using ResNet-50~\citep{he2016deep} (left), ViT~\citep{dosovitskiy2020image} (middle), and GPT-2~\citep{radford2019language} (right).} 
    \label{fig:imagenet}
\end{figure}
}

\rebuttal{
\section{Discussion} 
As demonstrated in the analysis of the Lyapunov function in Theorem~\ref{thm:main}, the Lion-$\phi$ dynamics exhibit a distinct nature when compared to typical momentum-based methods like Polyak, Nesterov momentum, and Hamiltonian descent, all of which can be conveniently understood as certain generalized dissipative Hamiltonian systems. While the Lyapunov function provides a powerful characterization of the dynamical behavior, our intuitive understanding of the Lion-$\phi$ dynamics remains obscured because we lack a ``physical intuition" or constructive derivation like the standard optimization algorithms. 
This invites more studies in studies and understandings in future works.  

The connection between Lion-$\phi$ and Nesterov momentum and accelerated mirror descent suggests the possibility of acceleration phenomena in variants of Lion-$\phi$, which opens an exciting avenue for future exploration and research. It might be possible to find novel accelerated algorithms based on the Lion-$\phi$ family.

It is surprising and compelling that an algorithm found by a random search program has such a rich and intriguing theoretical basis. The reasons for this remain elusive, whether it is a coincidence or due to some inherent necessity. For instance, the design of the search space in \citet{chen_symbolic_2023} may in some way entails a high likelihood of discovering theoretically sound algorithms with random search. Understanding the underlying logic here could lead to future advancements in automatic machine-based algorithm discovery.

Regarding applications, since Lion-$\phi$ offers a broader family than Lion, it is possible to find within the Lion-$\phi$ family new algorithms that outperform Lion in various tasks and metrics. Additionally, by using different values of $\phi$, Lion-$\phi$ can be utilized to address different types of constraint optimization problems.
} 

\section{Acknowledgement}
The research is conducted in Statistics \& AI group at UT Austin, which receives supports in part from NSF CAREER1846421, SenSE2037267, Office of Navy Research, and NSF AI Institute for Foundations of Machine Learning (IFML).

\bibliographystyle{plainnat}
\bibliography{iclr2024_conference}

\newpage \clearpage 
\appendix 

\section{Examples of $\phi$} 
\label{sec:Kexamples}
We provide a list of examples of $\phi$ and the corresponding $\dd \phi$ and $\phi^*$. 
It is useful to define the following indicator functions of set $\{z=0\}$: 
\bb 
\delta(z) = \begin{cases}
0 & \text{if $z =0$}  \\
+\infty & \text{if $z \neq 0$}. 
\end{cases}, && 
\ind(z) = \begin{cases}
0 & \text{if $z =0$}  \\
1 & \text{if $z \neq 0$}. 
\end{cases},
\ee 
Note that $\delta$ is the conjugate function of $f(x)=x$, as $\delta(x) = \sup_{z} x\tt z$. 
\paragraph{$\ell_p$ norm} 
When $\phi(x) = \norm{x}_p = (\sum_i \abs{x_i}^p)^{1/p}$ for $p \geq 1$, we can take 
$$
\dd \phi(x) = \frac{ \sign(x) \abs{x}^{p-1}}{\norm{x}_p^{p-1}}, 
$$  and 
$$
\phi^*(x) = \sup_{z} x \tt z - \norm{z}_p
= \sup_{c\geq 0} \norm{x}_q c - c  = \delta(\norm{x}_q \leq 1), 
$$
where $q$ is the conjugate number of $p$, satisfying $\frac{1}{p} + \frac{1}{q} = 1$. 
Hence, Lion-$\phi$ with $\ell_p$ norm
correspond to solving 
$$
\min_x f(x) ~~~s.t.~~~~ \norm{x}_q \leq 1/\lambda. 
$$

\paragraph{Group $\ell_p$ norm}
Assume $x$ is partitioned into a number of groups: $x = [x_{\mathcal G_i}]_{i=1}^k$. 
Consider the group $\ell_p$ norm:  
$\phi(x) = \sum_{i=1}^k \norm{x_{\mathcal G_i}}_p$. 
Then, we can take 
$$
\dd \phi(x) = \left [ \frac{ \sign(x_{\mathcal G_i}) \abs{x_{\mathcal G_i}}^{p-1}}{\norm{x_{\mathcal G_i}}_p^{p-1}}  \right ]_{i=1}^k
$$
The conjugate function is 
$$
\phi^*(x) = \sup_{z} \sum_{i=1}^k x_{\mathcal G_i} \tt z_{\mathcal G_i}  - \norm{z_{\mathcal G_i}}_p
= \sum_{i=1}^k \mathbf{\delta}( \norm{x_{\mathcal{G}_i}}_q \leq 1 ). 
$$
Hence, Lion-$\phi$ with grouped $\ell_p$ norm corresponds to solving  
$$
\min_x f(x)~~~~s.t.~~~~ \norm{x_{\mathcal G_i}}_q \leq 1/\lambda, ~~~~\forall i. 
$$

\paragraph{Lower Truncated $\ell_1$ Norm}
Consider 
$\phi(x) = \sum_{i=1}^d \max(\abs{x_i} - e, 0)$ where $e >0$. We can take  
\bbb \label{equ:indeps}
\dd\phi(x) = \ind(\abs{x} \geq  e)  \sign(x), 
\eee  
which uses $\sign(x)$ as Lion, but zeros out the gradient on the elements with absolute values smaller than $e $. 
The conjugate is  
\bb 
\phi^*(x) 
& = \sup_z  \sum_{i=1}^d (x_i z_i  - \max(\abs{z}_i - e, 0)) \\ 
& = \sup_{z, c} \sum_{i=1}^d (x_i z_i - c_i)~~~~s.t.~~~~ c_i\geq 0, ~~~  c \geq \abs{z_i} -  e \\
& = \sup_{c\geq 0}
\sum_{i=1}^d \abs{x_i} (c_i + e) - c_i \\ 
&= \sum_{i=1}^d \delta(\abs{x_i} \leq 1) +  e \abs{x_i} \\
& = \delta(\norm{x}_\infty \leq 1) + e \norm{x}_1. 
\ee 
Hence, Lion-$\phi$ 
corresponds to solving 
\bbb \label{equ:fgx}
\min_x \alpha f(x) + e \gamma \norm{x}_1 ~~~~~s.t.~~~~~ 
\norm{x}_{\infty}\leq 1/\lambda.  
\eee 
Hence, truncating the small gradients in Lion induces an $\ell_1$ penalty,
which encourages the sparsity of the final solution. 

\paragraph{Lower (Vector-wise) Truncated $\ell_p$ Norm}
Consider 
$\phi(x) = \max(\norm{x}_p - e, 0)$. We have 
\bb 
\dd\phi(x) = \ind(\norm{x}_p - e \geq 0) \frac{\sign(x)\abs{x}^{p-1}}{\norm{x}_p^{p-1}},
\ee 
in which the gradient is zeroed out when $\norm{x}_p \leq e$. 
The conjugate is  
\bb 
\phi^*(x) 
& = \sup_z  (x \tt z  - \max(\norm{z}_p - e, 0)) \\ 
& = \sup_{z, c} (x\tt z - c)~~~~s.t.~~~~ c \geq 0, ~~~  c \geq \norm{z}_p -  e \\
& = \sup_{c\geq 0}\norm{x}_q(c + e ) - c \\
&= \delta(\norm{x}_q \leq 1) +  e \norm{x}_q.  
\ee 
Hence, Lion-$\phi$ 
corresponds to solving 
$$
\min_x \alpha f(x) + e \gamma \norm{x}_q ~~~~~s.t.~~~~~ \norm{x}_q\leq 1/\lambda. 
$$

\paragraph{Sorting Norm}
For $x = [x_1, \ldots, x_d]$,
let $\abs{x_{(1)}} \geq \abs{x_{(2)}} \ldots$ be the sorting of the elements by absolute values. Define 
\bb 
\text{Sorting norm:}~~~~~
\phi(x) = \sum_i c_i \abs{x_{(i)}},
\ee 
where $c_1 \geq c_2 \geq ...\geq 0$ is a descending non-negative sequence.
The sorting norm 
is convex because it can be represented as the supreme of a set of convex functions, by  the rearrangement inequality, as follows  
\bb 
\phi(x) = \max_{\sigma \in \Gamma} \sum_{i=1}^d c_{\sigma(i)} \abs{x_i},  
\ee 
where $\Gamma$ denotes the set of permutations on $\{1,\ldots, n\}$. One subgradient of $\phi$ is 
\bb 
\dd \phi(x)_i = c_{rank(i, x)} \sign(x_i), 
\ee 
where $rank(i,x)$ denotes the rank of $\abs{x_i}$ in $x$. 
\bb 
\phi^*(x) 
& = \sup_{z} \left \{ x \tt z - \sum_i c_i \abs{z_{(i)}}  \right \} \\ 
& = \sup_{z \geq 0}
\left \{ 
\sum_i\abs{x_{(i)}} \times z_{(i)} - \sum_i c_i z_{(i)}
\right \} \ant{by rearrangement inequality}  \\
& = \sup_{w \geq 0}
\left \{ 
\sum_i(\abs{x_{(i)}} - c_i) \times 
(\sum_{j\geq i} w_j) 
\right \} \ant{let $z_{(i)} = \sum_{j\geq i} w_j$, $w_j\geq 0$}  \\
& = \sup_{w \geq 0}
\left \{ 
\sum_j \sum_{i\leq j}(\abs{x_{(i)}} - c_i) \times 
w_j 
\right \} \ant{let $z_{(i)} = \sum_{j\geq i} w_j$, $w_j\geq 0$}  \\
& = \sum_j \delta (\sum_{i\leq j} \abs{x_{(i)}} \leq \sum_{j\leq i} c_j) 
\ee 
Hence, Lion-$\phi$ corresponds to imposing a sequence of bounds on the cumsum of the sorted $x$: 
\bb 
\min_x f(x) ~~~s.t.~~~ 
\sum_{j\leq i} \abs{x_{(i)}} \leq C_i,
~~~~~\text{where} ~~ C_i = \sum_{j\leq i} c_j. 
\ee  

An interesting special case is when $c_i = \ind(i \leq i^{cut})$ for some integer $i^{cut} \in\{1,\ldots, d\}$, so that 
\bb 
\phi(x) = \sum_{i\leq i^{cut}} \abs{x_{(i)}}, &&
\dd \phi(x) = \ind(\abs{x} \geq x_{(i^{cut})} )\sign(x),  
 \ee 
 in which we zero out the updates of the elements whose absolute values are smaller than the $i^{cut}$-th largest element. 
 It is useful to compare this with 
 \eqref{equ:indeps} which applies the truncation based on a fixed number $\epsilon$, rather than the percentile. 
 
 The conjugate is 
 $$
\phi^*(x) = \sum_{j\leq i^{cut}} \delta( \abs{x_{(j)}} \leq 1) 
+ 
\delta( 
\norm{x}_1 \leq i^{cut} 
)
$$
Then, Lion-$\phi$ in this case corresponds to solving 
\bb
\min_x f(x) ~~~s.t.~~~
\norm{x}_1 \leq i^{cut}/\lambda, ~~~~~ 
\norm{x}_\infty \leq 1/\lambda, 
\ee 
in which the percentile-based truncation effectively imposes a constraint on the $\ell_1$ norm of $x$. It is different from \eqref{equ:fgx} 
in which the $\ell_1$ norm appears as a regularization term in the objective, rather than as a hard constraint. 

\paragraph{Entropy}
Consider 
$\phi(x) = \sum_{i=1}^d \frac{1}{a}\log \left ( \frac{1}{2}(\exp(a x_i) +  \exp(-a x_i)) \right )$, where $a >0$. We have 
$$\dd\phi(x) = \frac{\exp(a x) - \exp(-a x)}{\exp(a x) + \exp(-a x)} = \tanh(ax).$$
Taking the inverse, we have 
$\dd\phi^*(x) = \frac{1}{2a} \log \frac{1+x}{1-x}$, with domain in $\norm{x}_\infty \leq 1$. 
by integration, the conjugate function is hence,
\bb 
\phi^*(x) 
& =
\sum_{i=1}^d \frac{1}{2a} (x_i+1)\log(x_i+1) 
+ \frac{1}{2a} (1-x_i) \log (1-x_i) 
 + \delta(\norm{x}_\infty <1). 
\ee 
Lion-$\phi$ correspond to solving an entropy-regularized optimization: 
$$
\min_{x} \alpha f(x) +  \frac{\gamma }{\lambda }E(\lambda x)~~~~s.t.~~~~ \norm{x}_\infty \leq 1/\lambda,  
$$
where $E(x) = \sum_{i=1}^d \frac{1}{2a} (x_i+1)\log(x_i+1) 
n+ \frac{1}{2a} (1-x_i) \log (1-x_i)$. 
\paragraph{Huber Loss}
For $a \geq  0$, define the Huber loss: 
\bb 
\phi(x) = \sum_{i=1}^d \text{Huber}_a(x_i)
&&\text{where } &&
\text{Huber}_a(x_i) = 
\ind(\abs{x_i}\geq a) \times  \abs{x_i} + 
\ind(\abs{x_i} < a) \times \frac{1}{2 a} x_i^2, 
\ee 
We have 
\bb  
\dd\phi(x) = \mathrm{Clip}(x, -a, a)/a, && \text{with} &&
\mathrm{Clip}(x_i, a, b)  
= \begin{cases}
x_i & \text{if  $x \in [a,b]$} \\  
b & \text{if $x >b$} \\
a & \text{if $x < a$} .
\end{cases}
\ee 
The conjugate is 
$$
\phi^*(x) = \frac{a}{2} \norm{x}^2_2 + \delta(\norm{x}_\infty \leq 1), 
$$
\bb 
\phi^*(x)
& = 
\sum_{i=1}^d \max( \sup_{\abs{z} \geq a} x_i z_i - \abs{z_i} ,
~~~ \sup_{\abs{z_i} < a} x_i z_i - \frac{1}{2a} z_i^2 ) \\  
& = 
\sum_{i=1}^d \max 
\left ( \delta(\abs{x_i} \leq 1) + a (\abs{x_i}-1), ~~
\frac{1}{2} a x_i^2 
\right) \\
& = \sum_{i=1}^d \delta(\abs{x_i} \leq 1) + \frac{1}{2} a x_i^2 \\
& =\frac{a}{2} \norm{x}_2^2 +  \delta(\norm{x}_\infty\leq1). 
\ee 

\paragraph{Relativistic}
Consider 
$\phi(x)  = \sum_{i=1}^d \sqrt{x_i^2 + e^2}$, 
then $\dd \phi(x) = \frac{x}{\sqrt{x^2 + e^2}}$, and 
\bb 
\phi^*(x) 
& = \sup_z \left (\sum_{i=1}^d x_i z_i - \sqrt{z_i^2 + e^2} \right ) \\ 
& =\sum_{i=1}^d  \frac{x_i^2 e}{\sqrt{1-x_i^2}}  
-  \frac{e}{\sqrt{1 - x_i^2}} 
\ant{Solution: $z_i^2 =  \frac{x_i^2 e^2}{1 - x_i^2}$}
\\ 
& =\sum_{i=1}^d  - e \sqrt{1-x_i^2} + \delta(\abs{x_i} \leq 1) \\ 
& = \sum_{i=1}^d - e \sqrt{1-x_i^2} + \delta(\norm{x}_\infty\leq1). 
\ee 

A related case is
\bb 
\phi(x) =  \abs{x} - e \log (\abs{x}/e + 1), &&\text{with} &&
\dd\phi(x) = \frac{x}{\abs{x} + e},
\ee 
whose conjugate function is 
\bb 
\phi^*(x) 
& = \sup_z \left ( \sum_{i=1}^d x_i z_i
-  \abs{z_i} + e \log (\abs{z_i}/e+1) \right ) \\
& =  
\sum_{i=1}^d \abs{x_i}^2 e/(1-\abs{x_i}) - 
\abs{x_i}e/(1-\abs{x_i}) 
+ e \log (1/(1-\abs{x_i})) 
\ant{Solution: $z = \abs{x}e/(1-\abs{x})$} \\
& =   \sum_{i=1}^d - e(\abs{x_i} + \log (1-\abs{x_i})) + \delta(\norm{x}_\infty < 1).
\ee

\section{Proofs}

\subsection{Convex Function Preliminaries} 
\label{sec:convexity}

\paragraph{Lemma~\ref{lem:key00}} 
\emph{ 
Assume $\phi, \phi^*$ is a closed convex conjugate pair and $\dd\phi$, $\dd\phi^*$ are their subgradients, we have 
\bbb \label{equ:keyineqs22} 
(\dd\phi(x) - \dd\phi(y))\tt (x-y) \geq 0, 
&&
(\dd\phi(x) - y) \tt 
(x - \dd\phi^*(y)) \geq 0. 
\eee  
} 
\begin{proof}
1) By definition of subgradient, we have 
\bb 
\phi(y) - \phi(x) \geq \dd \phi(x)\tt (y - x) \\ 
\phi(x) - \phi(y) \geq \dd \phi(y)\tt (x - y). 
\ee 
Summing them together yields $(\dd\phi(x) - \dd\phi(y))\tt (x-y) \geq 0.$

2) Because  $\dd\phi^*(y) \in \partial \phi^*(y)$, we have 
$$
\phi^*(\dd\phi(x)) -\phi^*(y) 
\geq \dd\phi^*(y) \tt (\dd\phi(x) - y),  
$$
Because 
$\dd\phi(x) \in \partial \phi(x)$, 
by the property of conjugate functions, 
we have 
$x \in \partial \phi^*(\dd\phi(x))$, and hence 
$$
\phi^*(y)  - \phi^*(\dd\phi(x)) 
\geq x  \tt (y  - \dd\phi(x)). 
$$
Summing the two inequalities above yields 
\bb 
(\dd\phi(x) - y)\tt ( \dd\phi^*(y) - x)  
\leq 
 (\phi^*(\dd\phi(x)) -\phi^*(y) ) + (\phi^*(y) - \phi^*(\dd\phi(x))) 
= 0.
\ee
\end{proof}

\subsection{Connection with Nesterov Momentum} 
\label{sec:nesterov}
%

\begin{lem} 
The Lion-$\phi$ ODE is 
\bb
& \dot x_t = \dd\phi(m_t - \varepsilon (\alpha \dd f(x_t) + \gamma m_t)) - \lambda x_t \\ 
& \dot m_t =  - \alpha \dd f(x_t) - \gamma m_t . 
\ee 
is equivalent to 
\bbb \label{equ:lionodexonly} 
\dd^2\phi^*(\dot x_t + \lambda x_t) (\ddot x_t + \lambda \dot x_t) + \varepsilon \alpha \dd^2f(x_t) \dot x_t + 
 \gamma \dd\phi^*(\dot x_t + \lambda x_t) + 
\alpha \dd f(x_t)  = 0, 
\eee  
if $\phi^*$ and $f$ are twice differentiable. 

In particular, if $\phi(x) = \norm{x}_2^2/2$, we have 
\bbb \label{equ:odephi2} 
\ddot x_t + (\lambda + \gamma) \dot x_t + \varepsilon \alpha \dd^2f(x_t) \dot x_t + 
  \gamma \lambda x_t + 
\alpha \dd f(x_t)  = 0. 
\eee 
This ODE minimizes $F(x) = \alpha f(x) + \gamma\lambda \norm{x}^2_2/2$. 
\end{lem}
\paragraph{Remark} We have the following observations from \eqref{equ:odephi2}: 

1) The role of the weight decay $\lambda$
and momentum damping coefficient $\gamma$ is symmetric in \eqref{equ:odephi2}. 

2) 
When either the weight decay or momentum damping is turned off, i.e., 
$\gamma \lambda = 0$,
the $\ell_2$ regularization in $F(x)$ is turned off, and 
we have 
\bbb \label{equ:odephi3} 
\ddot x_t + (\lambda + \gamma) \dot x_t + \varepsilon \alpha \dd^2f(x_t) \dot x_t + 
\alpha \dd f(x_t)  = 0, 
\eee 
which coincides with the \emph{high-resolution} ODE 
\citep{shi2021understanding}
that serves as a continuous-time modeling of Nesterov momentum for minimizing $f(x)$.

3) 
The Hessian-dependent damping term $\dd^2 f(x_t) \dot x_t$
arises to due the gradient enhancement ($\varepsilon>0$), and it is known to play a key role in Nesterov momentum and acceleration  \citep{attouch2016fast, shi2021understanding}. When we turn off the gradient enhancement ($\varepsilon=0$), we get 
\bb 
\ddot x_t + (\lambda +\gamma) \dot x_t + \alpha \dd f(x_t) =0,
\ee 
which is the ODE 
for Polayk momentum, 
the equation of motion 
of a  ball with unit mass moving in a potential field $\alpha f(x)$ 
with a friction coefficient $(\lambda + \gamma)$. 


\begin{proof} 
We want to cancel out $m_t$. 
The first equation yields  
\bbb \label{equ:m}
(1-\varepsilon \gamma) m_t = 
\left ( \dd\phi^*(\dot x_t + \lambda x_t) + \varepsilon \alpha \dd f(x_t)   \right ). 
\eee 
Plugging it into the second equation yields 
\bbb \label{equ:dotm} 
\begin{split}
(1-\varepsilon \gamma) \dot m_t 
& = - \alpha (1-\varepsilon \gamma) \dd f(x_t) - {\gamma}\left ( \dd\phi^*(\dot x_t + \lambda x_t) + \varepsilon \alpha \dd f(x_t)   \right ) \\ 
& = - {\alpha }
\dd f(x_t) - {\gamma} \dd\phi^*(\dot x_t + \lambda x_t). 
\end{split}
\eee  
Combining \eqref{equ:m} and \eqref{equ:dotm} yields
\bb 
\ddt \left ( \dd\phi^*(\dot x_t + \lambda x_t) + \varepsilon \alpha \dd f(x_t)   \right ) = - {\alpha }
\dd f(x_t) - {\gamma} \dd\phi^*(\dot x_t + \lambda x_t).  
\ee  
Or 
\bb 
\dd^2\phi^*(\dot x_t + \lambda x_t) (\ddot x_t + \lambda \dot x_t) + \varepsilon \alpha \dd^2f(x_t) \dot x_t + 
 \gamma \dd\phi^*(\dot x_t + \lambda x_t) + 
\alpha \dd f(x_t)  = 0. 
\ee 
\end{proof}

\subsection{Discrete-time Schemes of Lion-$\phi$}
In the  most general form, the Euler approximation of the Lion-$\phi$ ODE with step size $\epsilon$ is 
\bbb \label{equ:alldisc} 
\begin{split}
& x_{t+1} = x_t + \epsilon (\dd \phi(m_t - \varepsilon (\alpha \dd f(x_t) + \gamma m_t )) - \lambda x_t ) \\ 
& m_{t+1} = m_t - \epsilon (\alpha \dd f(x_t)+ \gamma m_t) ,
\end{split} 
\eee  
The discrete Lion-$\phi$ scheme  in \eqref{equ:lionK} 
is recovered when $\alpha = \gamma$, 
$\beta_1 = 1-\varepsilon \gamma$, $\beta_2 = 1-\epsilon \gamma$. 
By scaling $f(x)$ by a positive multiplicative ratio, \eqref{equ:lionK} in fact covers all cases of \eqref{equ:alldisc} when $\gamma \neq 0$. 

When $\gamma = 0$, however, \eqref{equ:alldisc} reduces to a momentum-undamped variant of Lion-$\phi$: 
\bb \text{Undamped Lion-$\phi$:}~~~~~~~
\begin{split}
& x_{t+1} = x_t + \epsilon (\dd \phi(m_t - \beta_1 \dd f(x_t)) - \lambda x_t ) \\ 
& m_{t+1} = m_t - \beta_2 \dd f(x_t),
\end{split}
\ee 
which is the Euler approximation of Lion-$\phi$ ODE  $\gamma = 0$, step size $\epsilon $, and $\beta_1 = \varepsilon \alpha$, and $\beta_2 = \epsilon \alpha$. Due to $\gamma = 0$, the undamped Lion-$\phi$ amounts to solving $\min_x f(x)$, without the regularization $\phi^*(\lambda x)$.

The connection to Polyak and Nesterov momentum 
discussed in Section~\ref{sec:nesterov} extends to discrete-time forms. 
From the first equation \eqref{equ:alldisc}, we have 
\bb 
m_t = \frac{1}{1-\varepsilon \gamma }\left ( \dd\phi^*\left (\frac{x_{t+1} - x_t}{\epsilon} + \lambda x_t \right ) + \varepsilon \alpha \dd f(x_t) \right). 
\ee 
Plugging it into the second equation of \eqref{equ:alldisc}, we get 
{\scriptsize   
\bb 
\left ( \dd\phi^*\left (\frac{x_{t+2} - x_{t+1}}{\epsilon} + \lambda x_{t+1} \right ) + \varepsilon \alpha \dd f(x_{t+1}) \right)  = (1-\epsilon \gamma ) \left ( \dd\phi^*\left (\frac{x_{t+1} - x_t}{\epsilon} + \lambda x_t \right ) + \varepsilon \alpha \dd f(x_t) \right)  
- (1-\varepsilon \gamma)\epsilon \alpha \dd f(x_t). 
\ee 
} 
Hence, 
$$
\dd\phi^*\left (\frac{x_{t+2} - x_{t+1}}{\epsilon} + \lambda x_{t+1} \right ) 
= - \varepsilon \alpha \dd f(x_{t+1})
+  (1-\epsilon \gamma ) \dd\phi^*\left (\frac{x_{t+1} - x_t}{\epsilon} + \lambda x_t  \right ) 
+ (\varepsilon - \epsilon) \alpha  \dd f(x_t).   
$$

When $\dd\phi^*(x) = x$,  we have  
$$x_{t+2} = (1- \epsilon \lambda )x_{t+1}  - \epsilon  \varepsilon \alpha \dd f(x_{t+1})+  (1-\epsilon \gamma ) (({x_{t+1} - x_t})  + \epsilon \lambda x_t  ) + \epsilon (\varepsilon - \epsilon) \alpha  \dd f(x_t).   
$$
It is simplified into  
$$
x_{t+2} = (1- \epsilon^2 \lambda \gamma) x_{t+1} 
 - \epsilon^2 \alpha \dd f(x_{t+1}) 
 + 
  (1-\epsilon \gamma ) (1- \epsilon \lambda ) (x_{t+1} - x_t) - 
  \epsilon (\varepsilon - \epsilon) \alpha  ( \dd f(x_{t+1})  -  \dd f(x_t)).   
$$

When $\varepsilon > \epsilon$ (corresponding to $\beta_1 <\beta_2$ in Lion-$\phi$ \eqref{equ:lionK}),  
this can be shown to be identical to the Nesterov momentum 
algorithm for minimizing $F(x) =\alpha f(x) +  \lambda \gamma \norm{x}^2_2/2$.  When $\varepsilon = \epsilon$ (corresponding to $\beta_1=\beta_2$ in \eqref{equ:lionK}), it is identical to Polyak momentum. 

\subsection{Frank-Wolfe and Mirror Descent} 
\label{sec:fw}
\paragraph{Frank-Wolfe}
When $\varepsilon\gamma = 1$, Lion-$\phi$  reduces to 
\bbb \label{equ:fwdd}
\dot x_t = \dd \phi(-\dd f(x_t)) - \lambda x_t, 
\eee 
where we also set $\varepsilon \alpha = 1$ without loss of generality. 
In this case, the ODE monotonically decreases the objective 
$$ F(x) = f(x) + \frac{1}{\lambda}\phi^*(\lambda x),$$  without resorting to an additional Lyapunov function.  
This can be seen from  
\bb
\ddt F(x_t) = 
(\dd f(x) +\dd \phi^*( \lambda x ))\tt (\dd\phi(-\dd f(x)) -  \lambda x) \leq 0,
\ee
where the inequality follows Lemma~\ref{lem:key00}. 


The Euler discretization of \eqref{equ:fwdd} is 
\bbb   \label{equ:fwdddisc}
x_{t+1}  = x_t + \epsilon \left(\dd \phi(-\dd f(x_t)) - \lambda x_t \right ). 
\eee
This can also be derived from conditional gradient descent,  or Frank–Wolfe. 
To see this, recall that the conditional gradient descent update for the $F(x)$ above is
\bb 
y_{t+1} & = \argmin_{x}\left \{  \dd f(x_t)\tt (x - x_t) + \frac{1}{\lambda}\phi^*(\lambda x)  \right\} \\
x_{t+1} & = x_t + \epsilon_0( y_{t+1}- x_t),
\ee 

Solving $y_{t+1}$ yields 
\bb 
y_{t+1} = \frac{1}{\lambda} \dd \phi( -\dd f(x_t) ),&&\text{and hence} &&
x_{t+1} = (1-\epsilon_0 ) x_t + \frac{\epsilon_0}{\lambda} \dd \phi(-\dd f(x_t)). 
\ee 
Taking $\epsilon = \epsilon_0 \lambda$ yields \eqref{equ:fwdddisc}.


\paragraph{Dual Space Preconditioning and 
Mirror Descent}
When we further set $\lambda= 0$ in \eqref{equ:fwdddisc}, 
Lion-$\phi$  reduces to 
\bbb \label{equ:mirrordescent}
x_{t+1}  = x_t + \epsilon \dd \phi(-\dd f(x_t)), 
\eee 
When $\dd\phi(0) = 0$, 
Eq.~\eqref{equ:mirrordescent} 
is dual space preconditioning \citep{maddison2021dual}, 
which is closely related to mirror descent \citep{nemirovskij1983problem}, for minimizing $f(x)$. To see the connection with mirror descent, 
note that \eqref{equ:mirrordescent} is equivalent to 
\bb 
x_{t+1} = x_t + \epsilon \delta_t, 
&& \text{with} &&
\delta_t = \argmin_\delta \left\{  \dd f(x_t) \tt \delta + \phi^*(\delta) \right\}.  
\ee 
Because $\phi^*$ and $\phi$ are differentiable, then $\dd \phi(0)=0$ implies $\dd\phi^*(0)=0$, and hence $\phi^*$ achieves the minimum at zero. 
In this case, 
$\phi^*(\delta) - \phi^*(0)$ can be viewed as a Bregman divergence, and hence justifying  the connection of \eqref{equ:mirrordescent} with mirror descent.
Recall that the Bregman divergence $B_{h}(x ~||~ y)$ is the Bregman divergence associated with a convex function $h \colon \RR^d \to \RR$ is defined as 
$$
B_{h}(x ~||~ y) = h(x) - h(y) - \dd h(y) \tt (x- y). 
$$
With $\dd\phi^*(0)=0$, 
it is then easy to show 
\bb 
\phi^*(\delta) - \phi^*(0)
= B_{\phi^*}(\delta ~||~0) 
= B_{\phi^*_t} (x_t + \epsilon \delta ~||~x_t), 
\ee 
where $\phi^*_t = \phi^*\left (\frac{x - x_t}{\epsilon} \right )$.

\subsection{Lion-$\phi$ without gradient Enhancement ($\varepsilon =0$)}
\label{sec:nogradientcorrect}

\begin{thm}\label{thm:lionnongrad1}
Consider the ODE of Lion-$\phi$-W without gradient correction: 
\bbb \label{equ:lionnong} 
\begin{split}
& \dot x_t = \dd\phi(m_t) -  \lambda x_t \\
& \dot m_t = -   \alpha \dd f(x_t) - \gamma m_t, 
\end{split}
\eee  
with $\lambda, \alpha, \gamma > 0$. 
Its fixed point is the minimum of 
$$
\min_{x}  \alpha f(x) + \frac{\gamma}{ \lambda} \phi^*(\lambda x). 
$$
It yields the following Lyapunov function: 
\bb 
H(x, m) = 
 {\alpha} f(x) +  \frac{\gamma}{\lambda}\phi^*(\lambda x) + 
 (\phi^*(\lambda x) + \phi(m) - \lambda x \tt m). 
\ee 
\end{thm}

\begin{proof} 
Observe that 
\bb 
& \dd_x H(x, m) = \alpha \dd f(x) 
+ ({\gamma} +\lambda )\dd\phi^*(\lambda x) - \lambda  m  \\
& \dd_m H(x, m) = \dd\phi(m) -\lambda x, 
\ee 
and \eqref{equ:lionnong} can be written into 
\bb 
&  \dot x_t \defeq  V_x(x_t, m_t) 
= \dd_m H(x_t, m_t) \\ 
& \dot m_t \defeq V_m(x_t, m_t)
=  - \dd_x H(x_t, m_t) -  \hat H_m(x_t, m_t), 
\ee 
with $\hat H_m(x_t, m_t) = (\gamma +\lambda) (m_t - \dd\phi^*(\lambda x_t)).$ 
By Lemma~\ref{lem:key00}, we have 
$$
\hat H_m \tt (\dd_m H) 
= (m - \dd\phi^*(\lambda x))\tt (\dd\phi(m)-\lambda x) 
\geq 0. 
$$
Then 
\bb 
\ddt H(x_t, m_t ) & = 
\dd_x H \tt V_x 
+ \dd_m H\tt V_m \\
& = \dd_x H \tt  (\dd_m H)
+ \dd_m H \tt (- \dd_x H - \hat H_m) 
= - \dd_m H \tt \hat H_m \leq 0.
\ee 
In fact, this ODE has a Hamiltonian + descent structure \citep{maddison2018hamiltonian},  
as it can viewed as a Hamiltonian system damped with a descending force: 
\bb 
\begin{bmatrix}
\dot x_t \\
\dot m_t
\end{bmatrix} 
& = 
\blue{\underbrace{\begin{bmatrix} \blue{+ \dd_m H(x_t, m_t)}\\ 
 \blue{-\dd_x H(x_t, m_t )} 
 \end{bmatrix} }_{\blue{\text{Hamiltonian}}}} 
 -  
 \med{\underbrace{\begin{bmatrix} 0 \\
 (\gamma + \lambda ) {(m_t - \dd\phi^*(\lambda x_t))} 
 \end{bmatrix}}_{\text{ Descent}}}, 
\ee 
where the Hamiltonian component is orthogonal to the gradient $[\dd_x H, \dd _m H]$ of $H(x,m)$ and preserves the total energy $H(x,m)$,
and the descent component introduces a damping like effect to decrease the energy $H(x,m)$. 
\end{proof}

\subsection{Lion-$\phi$ without Weight Decay -- A Hamiltonian + Descent Derivation}
\label{sec:lionsolved}
When the weight decay in Lion-$\phi$ is turned off ($\lambda =0$), 
there is an alternative way to analyze it that is amendable to the Hamiltonian + descent structure in \eqref{equ:HVdual}. 

Recall that the Lion-$\phi$ ODE is of the following form when $\lambda =0$: 
\bbb \label{equ:xmthha}
\begin{split}
& \dot x_t = \dd \phi(\tilde m_t),~~~~~~
 \tilde m_t = m_t - \varepsilon (\alpha \dd f(x_t) + \gamma m_t) \\ 
& \dot m_t = - \alpha \dd f(x_t) - \gamma m_t \\
\end{split} 
\eee  
Assume $\varepsilon \gamma <1$. 
Define $\tilde \phi(m) = 
\frac{1}{1-\varepsilon \gamma}\phi((1-\varepsilon \gamma) m)$, 
and 
the following Lyapunov function: 
\bbb \label{equ:thisH} 
H(x, m) = \alpha f(x) + \tilde \phi(m) = \alpha f(x)  + 
\frac{1}{1-\varepsilon \gamma}\phi((1-\varepsilon\gamma) m ). 
\eee 
Note that $\dd_x H(x,m) = \alpha \dd f(x)$ and $\dd_m H(x,m) = \dd \phi((1-\varepsilon )m )$. 
One can decompose \eqref{equ:xmthha} into the following Hamiltonian + descent decomposition: 
\bb
\begin{bmatrix}
\dot x_t \\
\dot m_t 
\end{bmatrix}  
= \blue{\underbrace{\begin{bmatrix}
+ \dd_m H(x_t, m_t) \\
- \dd_x H(x_t, m_t)
\end{bmatrix}}_{\text{Hamiltonian}}}    
- \med{\underbrace{\begin{bmatrix}
\dd  \phi(\tilde m_t^0) - \dd  \phi(\tilde  m_t) \\
\gamma m_t
\end{bmatrix}}_{\text{Descent}}}, 
\ee 
where we define 
$\tilde m_t^0= (1- \varepsilon \gamma) m_t$
and hence $\tilde m_t- \tilde m_t^0 = - \varepsilon\alpha \dd f(x_t).$ 

Using the monotonicity of subgradient (Lemma~\ref{lem:key00}), 
one can show that the second component in the decomposition above is a descent direction of $H(x,m)$ in \eqref{equ:thisH}:  

1) Let 
$\hat \dd_x H_t \defeq -\dd \phi(\tilde m_t^0) + \dd  \phi(\tilde m_t )$, then it 
is a descent direction of $H(x, m)$, because 
\bb 
\dd_x H(x_t,m_t) \tt \hat \dd _x H_t 
& = \alpha \dd f(x_t) \tt \hat \dd_x H_t  \\ 
 & =  - \frac{1}{\varepsilon}(\tilde m_{t}^0 - \tilde m_t) \tt  (\dd \phi(\tilde m_t^0) - \dd \phi(\tilde m_t)) \leq 0,
\ee 
where we used the monotonicity of $\dd \phi(\cdot)$. 

2) If $m = 0$ is the minimum of $\phi$, 
then $\hat \dd _m H_t \defeq  - \gamma m_t$ is a descent direction of $H(x,m)$ because, 
\bb  
\dd_m H(x_t,m_t) \tt \hat \dd_m H_t  
= - \gamma \dd  \phi((1-\varepsilon \gamma)m_t) \tt m_t  
\leq   \frac{\gamma}{1-\varepsilon \gamma}(\phi(0)  -  \phi((1-\varepsilon \gamma) m_t) ) \leq 0. 
\ee 
Hence, we have 

\bb 
\ddt H(x_t, m_t) 
& =  \dd_x H(x_t,m_t) \tt \hat \dd _x H_t + \dd_m H(x_t,m_t) \tt \hat \dd_m H_t  \\ 
& =  - \frac{1}{\varepsilon}(\tilde m_{t}^0 - \tilde m_t) \tt  (\dd \phi(\tilde m_t^0) -\dd \phi(\tilde m_t)) -\gamma \dd  \phi((1-\varepsilon \gamma)m_t) \tt m_t  
\leq 0. 
\ee 
Moreover, if $m=0$ is the unique minimum of $\phi$, and $\varepsilon \gamma < 1$, 
then $ \dd  \phi((1-\varepsilon \gamma)m_t) \tt m_t  =0$ implies that $m_t =0$, and one can show that the equilibrium points of \eqref{equ:xmthha} are stationary points of $H(x,m)$ using LaSalle's invariance principle.

\subsection{Main Result of Lion-$\phi$ ODE}
\label{sec:proofsmain}
\begin{thm}
Assume $\phi$ is convex with conjugate $\phi^*$. 
Assume $f, \phi, \phi^*$ are continuously differentiable. 
Assume $(x_t, m_t)$ is the solution of the following 
ODE:
\bb 
& \dot x_t = \dd\phi(\tilde m_t ) -\lambda  x_t , ~~~~~~~~\text{with} ~~~~~~ \tilde m_t = m_t - 
\varepsilon (\gamma  m_t + \alpha \dd f(x_t)), \\
& \dot m_t =  
- \alpha \dd f(x_t)  - \gamma  m_t,
\ee 
where $\alpha,\gamma,\lambda,\varepsilon > 0$ and $\epsilon \gamma \leq 1$. 
Let 
\bb 
H(x, m) = \alpha   f(x)   + 
\frac{ \gamma }{\lambda}
\phi^*(\lambda x) + 
\frac{1-\varepsilon \gamma }{1+ \varepsilon \lambda} 
(\phi^*(\lambda x) +  \phi(m)   - \lambda 
m\tt  x ).
\ee 
Then $H$ yields a Lyapunov function in that 
$$
- \ddt H(x_t, m_t) =  \Delta (x_t, m_t)
\defeq   \frac{\lambda +\gamma}{1+\varepsilon \lambda }
\Delta_1(x_t, \tilde m_t)
+  \frac{1-\varepsilon\gamma}{(1+\varepsilon \lambda ) }
\Delta_2(m_t, \tilde m_t)
 \geq 0,
$$
where 
\bb 
& \Delta_1(x, \tilde m ) = (\tilde m - \dd\phi^*(\lambda x))\tt 
(\dd\phi(\tilde m) - \lambda x), \\ 
& \Delta_2(m, \tilde m) = \frac{1}{\varepsilon} (\tilde m - m)  \tt (\dd\phi(\tilde m) - \dd\phi(m)). 
\ee 
Moreover, the accumulation points of all trajectories are stationary points of $F(x) = \alpha f(x) + \frac{\gamma}{\lambda }\phi^*(\lambda x)$. 
\end{thm} 

\begin{proof}
It is not obvious how to construct the Lyapunov function directly from the ODE. 
The following proof describes the process of discovering $H(x,m).$ 
We start by examing what inequalities we can write down 
using the monotonicity of $\dd \phi$ and $\dd \phi^*$ via Lemma~\ref{lem:key00}, 
and then work out the Lyapunov function backward. 

Write $\tilde m =  m - \varepsilon (\gamma m + \alpha \dd f(x))$.
Because $\dd \phi$ is a monotonic mapping, we have by Lemma~\ref{lem:key00} the following  key inequalities: 
\bb 
& (  - \tilde m +    \dd\phi^*(\lambda x))  \tt (\dd\phi(\tilde m) - \lambda x)  \leq 0,  \\ 
& (m - \tilde m)\tt (\dd\phi(\tilde m) - \dd\phi(m)) \leq 0, 
\ee 
or equivalently 
\bbb  
& (  \varepsilon \alpha  \dd f(x) - (1- \varepsilon \gamma )   m +   \dd\phi^*(\lambda x)) \tt  ( \dd\phi(\tilde m) - \lambda x) \leq 0  \label{equ:ineq11}\\ 
& \varepsilon ( \alpha \dd f(x)  + \gamma m  )\tt  ( (\dd \phi(\tilde m)-\lambda  x) - (\dd\phi(m)- \lambda x))  \leq 0 \label{equ:ineq22}
\eee  
Write $V_x = \dd\phi(\tilde m) - \lambda x$, and $V_m = - \alpha \dd f(x) - \gamma m$. So the ODE is $\dot x = V_x$ and $\dot m = V_m$. The inequalities can be rewritten into 
\bbb  
& (  \varepsilon \alpha  \dd f(x) - (1- \varepsilon \gamma )   m +   \dd\phi^*(\lambda x)) \tt  V_x  \leq 0  \label{equ:ineq11}\\ 
& - \varepsilon V_m \tt  ( V_x - (\dd\phi(m)- \lambda x))  \leq 0 \label{equ:ineq22}
\eee  

Taking  $\frac{1}{\varepsilon(1+\eta)}(Eq.~\eqref{equ:ineq11} + \eta \times  Eq.~\eqref{equ:ineq22})$ for any $\eta \geq 0$, we get 
\bb 
\left ( \alpha \dd f(x) - \frac{1  - \varepsilon \gamma (1 +  \eta) }{ \varepsilon (1 + \eta)} m  +  
\frac{1}{\varepsilon  (1 + \eta)}\dd\phi^*(\lambda x)  
\right ) \tt  V_x +  
\frac{\eta \varepsilon}{\varepsilon  (1 + \eta)}  ( \dd\phi(m)- \lambda x)) \tt V_m  \leq 0
\ee 
Define
$$
\tilde  H(x,m) 
= \alpha  f(x)    +\frac{1}{ \varepsilon (1+\eta )  \lambda } \phi^*(\lambda x) + \frac{1 - \varepsilon \gamma (1+ \eta )}{ \varepsilon (1+\eta ) }  \frac{1}{\lambda} \phi(m)   - 
\frac{1 - \varepsilon \gamma (1+ \eta )}{ \varepsilon (1+\eta ) }     m\tt  x.
$$
Then the inequality was reduced to 
\bb
\dd_x \tilde H(x, m)  \tt V_x 
+ 
 \frac{\varepsilon \eta \lambda}{1- \varepsilon \gamma (1+\eta)} 
\dd_m \tilde H(x, m)  \tt V_m 
 \leq 0.
\ee 
If we take $\eta$ such that 
\bbb \label{equ:etachoice} 
\frac{\varepsilon \eta \lambda}{1- \varepsilon \gamma (1+\eta)}  = 1,
\eee   then we have  when following $\dot x = V_x$ and $\dot m = V_m$, 
$$
\ddt \tilde H(x, m) 
= \dd_x \tilde H(x, m)  \tt V_x + 
\dd_m \tilde H(x, m)  \tt V_m  \leq 0. 
$$
Furthermore, when  \eqref{equ:etachoice} holds, we have 
\bbb 
   \eta = \frac{1-\varepsilon \gamma }{\varepsilon (\lambda + \gamma)} , 
  && 
    \frac{1}{\varepsilon (1+\eta )}= 
  \frac{ \lambda + \gamma}{1 + \varepsilon \lambda  },  
  &&
  \frac{1-\varepsilon \gamma (1+\eta)}{\varepsilon (1+\eta)} = \frac{(1-\varepsilon \gamma)\lambda }{1+ \varepsilon\lambda}, 
  \label{equ:etaeqs} 
\eee 
and hence  
\bb 
\tilde H(x,m)  
& =  
\alpha   f(x)    + 
\left (\frac{\lambda + \gamma}{(1+\epsilon \lambda )\lambda } 
- \frac{1-\varepsilon \gamma}{1+\varepsilon \lambda}  
\right )
\phi^*(\lambda x) +
\frac{1-\varepsilon \gamma}{1+\varepsilon \lambda} 
\left ( \phi^*(\lambda x) + 
\phi(m)   -     \lambda m\tt  x  \right ) \\ 
& = \alpha   f(x)   + 
\frac{\gamma}{\lambda } 
\phi^*(\lambda x) +
\frac{1-\varepsilon \gamma}{1+\varepsilon \lambda} 
\left ( \phi^*(\lambda x) + 
\phi(m)   -     \lambda m\tt  x  \right )  \\ 
& = H(x,m).  
\ee 
In this case, 
\bb 
& \ddt H(x,m) \\ 
& =  \frac{1}{\varepsilon(1+\eta)}(Eq.~\eqref{equ:ineq11} + \eta \times  Eq.~\eqref{equ:ineq22}) \\
& =  
\frac{\lambda + \gamma}{1+\varepsilon \lambda } 
\times Eq.~\eqref{equ:ineq11} 
+  \frac{1-\varepsilon \gamma}{\varepsilon (1+\varepsilon \lambda )}
\times  Eq.~\eqref{equ:ineq22}.  \ant{$     
    \frac{\eta}{\varepsilon (1+\eta )}= 
  \frac{ 1-\varepsilon \gamma }{\varepsilon (1 + \varepsilon \lambda )}$ from \eqref{equ:etaeqs}}
\\
& = - \frac{\lambda +\gamma}{1+\varepsilon \lambda }
(\tilde m - \dd\phi^*(\lambda x))\tt 
(\dd\phi(\tilde m) - \lambda x) 
- \frac{1-\epsilon\gamma}{(1+\epsilon \lambda)\varepsilon }
(\tilde m - m)  \tt (\dd\phi(\tilde m) - \dd\phi(m)) \leq 0. 
\ee 
To ensure that $\eta\geq 0$, we need $\varepsilon \gamma \leq 1$. 
\end{proof} 

\paragraph{LaSalle's invariance principle} 
Let $H(z)$ is a continuously differentiable Lyapunov function of $\ddt z_t = v(z_t)$, satisfying $\ddt H(z_t) \leq 0 $.  
By LaSalle’s Invariance Principle, 
the accumulation points of any trajectories of $\ddt z_t = v(z_t)$ is included in 
$$
\mathcal I = 
\{\text{the union of all trajectories $z_t$ 
satisfying $\ddt H(z_t) = 0$ for all $t\geq 0$ 
} \}. 
$$

For the Lion-$\phi$ ODE and its $H$, 
the points in $\mathcal I$ should satisfy  
$\tilde m_t = \dd \phi^*(\lambda x_t)$, 
which yields $\dd\phi(\tilde m_t) = \lambda x_t$, and hence 
$$
\dot x_t = \dd \phi(\tilde m_t) - \lambda x_t =0. 
$$
This suggests that $x_t$ 
is constant for the trajectories in  $\mathcal I$.  
Because $\tilde m_t = \dd\phi^*(\lambda x_t)$
and $\tilde m_t = m_t - \varepsilon (\alpha \dd f(x_t) + \gamma m_t) $, we have 
$$
(1-\varepsilon \gamma) m_t = \dd\phi^*(\lambda x_t) + \varepsilon \alpha \dd f(x_t)
$$
Hence, $(1-\varepsilon \gamma) m_t$ is also constant in the trajectories in $\mathcal I$. This suggests that $(1-\varepsilon \gamma)\dot m_t =0$ along the trajectories in $\mathcal I$, and hence 
\bb 
0&  = 
(1-\varepsilon \gamma) \dot m_t  \\ 
&= - (1-\varepsilon \gamma) (\alpha \dd f(x_t) + \gamma  m_t ) \\ 
& = - (1-\varepsilon \gamma)\alpha \dd f(x_t) 
- \gamma   \dd\phi^*(\lambda x_t) - \varepsilon  \gamma \alpha \dd f(x_t) \\
&  = - \alpha \dd f(x_t) - \gamma \dd \phi^*(\lambda x_t) \\
& = - \dd F(x_t)  \ant{$F(x) = \alpha f(x) + \frac{\gamma}{\lambda }  \phi^*(\lambda x)$}
\ee 
Hence, all trajectories in $\mathcal I$ are singleton points and are stationary points of the objective $F(x) = \alpha f(x) + \frac{\gamma}{\lambda }  \phi^*(\lambda x)$. 

\subsection{The Decomposition Structure} 
\label{sec:decompose}
We provide the decomposition structure \eqref{equ:HV} which provides a simplified proof of the Lyapunov property. 
\begin{lem}
For ODE 
$\dot x_t = V_x(x_t, m_t)$, $\dot m_t = V_m(x_t, m_t)$, let $H(x,m)$ be a function satisfying 
\bb 
\dd_x H(x,m) =  - \tilde  V_x(x,m)  + \eta V_m(x,m) \\
\dd_m H(x,m) = - \hat V_m (x,m)- \eta V_x(x,m) ,
\ee 
where $\eta \in \RR$ and $\hat V_x$ and $\hat V_m$ have positive inner products with $V_x$, $V_m$, respectively, that is, 
\bb \hat V_x(x,m) \tt 
V_x (x,m) \geq 0, && \hat V_m(x,m) \tt V_m(x,m) \geq 0,~~~~\forall x, m.
\ee 
Then we have 
$$
\ddt H(x_t, m_t) \leq 0. 
$$
\end{lem}
\begin{proof}
\bb 
\ddt H(x_t, m_t) 
& = \dd_x H\tt V_x + \dd_m H \tt V_m \\ 
& =  ( - \hat V_x + a V_m)\tt  V_x + (- \hat V_m - a V_x) \tt V_m \\ 
& = - (\hat V_x \tt V_x + \hat V_m \tt V_m) \leq 0. 
\ee 
 
\end{proof}

\begin{lem}\label{lem:decompH}
Under the condition of Theorem~\ref{thm:main}, 
let  
\bb 
& V_x(x,m) = \dd \phi(\tilde m) - \lambda x  \\ 
& V_m(x,m) = - \alpha\dd f(x) - \gamma m = \frac{\tilde m - m}{\varepsilon}
\ee  
and related 
\bb 
& \hat V_x(x,m) = \tilde m - \dd \phi^*(\lambda x)  = 
- \varepsilon \alpha \dd f(x) 
+ (1-\varepsilon \gamma) m 
- \dd \phi^*(\lambda x), 
\\   
& \hat V_m (x,m) =  \dd\phi(\tilde m) - \dd \phi(m).
\ee 
Then we have $\hat V_x \tt  V_x \geq 0$ and $\hat V_m \tt V_m \geq 0$ by Lemma~\ref{lem:key00}. 
Moreover, 
\bb
\dd_x H(x, m) =  - \eta'\hat V_x - \eta V_m \\ 
\dd_m H(x, m) =  - \eta \hat V_m + \eta V_x ,
\ee 
where $\eta = \frac{1-\varepsilon\gamma}{1+\varepsilon\lambda}$ 
and $\eta' = \frac{\gamma+\lambda }{1+\varepsilon\lambda}$.
This yields 
$$
\ddt H(x_t, m_t) = \dd_x H\tt V_x 
+ \dd _m H \tt V_m 
= - (\eta' \hat V_x\tt V_x + \eta \hat V_m \tt V_m) \leq 0. 
$$
\end{lem}
\begin{proof}

Let $\eta = \frac{1-\varepsilon \gamma}{1+\varepsilon \lambda } $. We have 
We have 
\bb 
\dd_m H(x, m)
& = \eta(\dd\phi(m) - \lambda x) \\ 
& = \eta (\dd\phi(\tilde m) - \lambda x + \dd\phi(m) - \dd\phi(\tilde m) )  \\
& =  \eta (V_x - \hat V_m). 
\ee

\bb
& \dd_x H(x,m)  \\
& = \alpha \dd f(x) + \gamma \dd \phi^*(\lambda x)  
+ 
\eta (\lambda \dd\phi^*(\lambda x) - \lambda m) \\
& = \alpha \dd f(x) + 
(\gamma + \eta \lambda)
 \dd\phi^*(\lambda x) - \eta \lambda m  \\
& =   
(\gamma+ \eta \lambda)
( \varepsilon \alpha \dd f(x) 
- (1-\varepsilon \gamma) m 
+ \dd \phi^*(\lambda x)) 
+  (\alpha - (\gamma+\eta\lambda)  \varepsilon \alpha)\dd f(x)
- (\eta\lambda  - (\gamma+\eta\lambda )(1-\varepsilon \gamma)) m \\ 
& =   
\frac{\gamma+\lambda}{1+\varepsilon\lambda} 
( \varepsilon \alpha \dd f(x) 
- (1-\varepsilon \gamma) m 
+ \dd \phi^*(\lambda x)) 
+ \eta \alpha \dd f(x)
+ \eta \gamma m  \\
& = - \frac{\gamma +\lambda }{1+\varepsilon\lambda } 
\hat V_x  - \eta V_m,
\ee
where we used the following identities on $\eta$: 
\bb 
& (\gamma+ \eta \lambda) = \gamma + 
 \frac{1-\varepsilon\gamma}{1+\varepsilon\lambda} \lambda = \frac{\gamma + \lambda }{1+\varepsilon \lambda} \\
& 1-(\gamma+\eta\lambda)\varepsilon = 1-\frac{\gamma+\lambda}{1+\varepsilon\lambda} \varepsilon = \frac{1-\varepsilon \gamma}{1+\varepsilon\lambda} = \eta  \\ 
& \eta \lambda - (\gamma+\eta \lambda )(1-\varepsilon\gamma)  = - \gamma +  \frac{\gamma+\lambda}{1+\varepsilon\lambda} \varepsilon \gamma  = \frac{\varepsilon\gamma^2-\gamma}{1+\varepsilon\lambda} = - \gamma \eta. 
\ee
\end{proof}

\subsection{Constraint Enforcing: Continuous Time}
\label{sec:constraintapp} 
When $\phi^*$ can possible take infinite values, the minimization of $H(x, m)$ becomes a constrained optimization. 
Let 
$\dom \phi^* = \{x \colon \phi^*(x) < +\infty\}$. 
The optimization can be framed as
$$
\min_{x, m} H(x, m) ~~~~~s.t.~~~~ \lambda x \in \dom\phi^*. 
$$
The Lion-$\phi$ algorithm would first steer $x_t$ to the region where $\phi^*$ has finite values, and then decrease the finite parts of the objective function. 
In the following, we show that  Lion-$\phi$ enforces the constraint with a fast linear rate: 
the distance from $\lambda x_t$ and $\dom \phi^*$ decays exponentially fast with time $t$, and once $\lambda x_{t_0} \in \dom \phi^*$, then $\lambda x_{t}$ stays within $\dom \phi^*$ for all $t  > t_0$.

\begin{thm}\label{thm:constraint}
Under the condition of Theorem~\ref{thm:main}, we have 
$$
\dist(\lambda x_t, \dom \phi^*) \leq \exp(\lambda (s-t)) ~\dist(\lambda x_s, \dom \phi^*). 
$$  
\end{thm}
\begin{proof}
Define $w_{s\to t} = \exp(\lambda (s-t))$. 
Integrating $\dot x_t = \dd\phi(\tilde m_t) - \lambda x_t$, we have 
\bb 
\lambda x_t =
(1- w_{s\to t})  z_{s\to t}  + w_{s\to t} (\lambda x_s), &&
\text{where}
&&
z_{s\to t} = \frac{\int_s^t w_{\tau\to t} \dd \phi(\tilde m_\tau ) \d \tau}{\int_s^t w_{\tau\to t} \d \tau},~~~~
\forall 0\leq s \leq  t. 
\ee 
We have $\dd\phi(\tilde m_\tau) \in \dom \phi^*$ from Lemma~\ref{lem:dom} and $\dom \phi^*$ is convex. Hence $z_{s\to t} $, as the convex combination of $\{\dd\phi(\tilde m_\tau)\}_\tau$, belongs to $\dom \phi^*$. 
For any $\epsilon>0$, let $\lambda \hat x_s \in \dom \phi^*$ to the point satisfying $\norm{\lambda \hat x_s - \lambda x_s} \leq \dist(\lambda x_s, \dom \phi^*) + \epsilon$. 
Hence, 
\bb 
\mathrm{dist}(\lambda x_t, ~ \dom \phi^*) 
& = \inf_{z \in \dom \phi^*} \norm{\lambda x_t  - z }  \\
& \leq \norm{\lambda x_t - (1-w_{s\to t}) z_{s\to t} - w_{s\to t} \lambda \hat x_s)} \\
& =  w_{s\to t} \norm{\lambda x_s - \lambda \hat x_s} \\
& \leq \exp(\lambda (s-t)) (\dist(\lambda x_s, \dom \phi^*) + \epsilon). 
\ee 
Taking $\epsilon \to 0$ yields 
$$
\dist(\lambda x_t, \dom \phi^*) \leq \exp(\lambda (s-t)) ~\dist(\lambda x_s, \dom \phi^*). 
$$
\end{proof}

\begin{lem}\label{lem:dom}
Assume $\phi$ is proper, closed and convex, 
and $\phi^*$ is the conjugate of $\phi$. 
We have 
$$
\partial \phi(z) \subseteq \dom \phi^*, ~~~~\forall z \in \dom \phi. 
$$
\end{lem} 
\begin{proof}
If $x\in \partial \phi(z)$, then $z$ attains the minimum of $\phi^*(x) = \sup_z \{x \tt z - \phi(z)\}$,
suggesting that $\phi^*(x) = x\tt z - \phi(z) <+\infty$, and hence $x \in \dom \phi^*$. 
\end{proof}

\subsection{Discrete Time Analysis}
\label{sec:discreteapp}
\begin{thm} \label{thm:discreteapp}
Assume $f\colon \RR^d\to \RR$ is $L$-smooth, and $\phi\colon \RR^d\to \RR$ is onvex.
Consider the following scheme: 
\bbb \label{equ:finiteupdate} 
\begin{split}
m_{t+1} & = \btwo m_t  - (1-\btwo) \dd f(x_t) \\ 
\tilde m_{t+1} & = \bone m_{t} - (1-\bone) \dd f(x_t) \\ 
x_{t+1} & = x_t + \lr
(\dd \phi(\tilde m_{t+1}) - \lambda x_{t+1}), 
\end{split} 
\eee  
where $\dd\phi$ is a subgradient of $\phi$, and 
$\bone,\btwo \in (0,1)$, and $\btwo>\bone$, and $\epsilon, \lambda > 0$. 
Let $\phi^*$ be the conjugate function of $\phi$. 
Define the following Lyapunov function: 
$$
H(x, m ) = f(x) + \frac{1}{\lambda } \phi^*(\lambda x) + \frac{\bone}{\epsilon \lambda  (1-\bone) + (1-\btwo)}(\phi^*(\lambda x) + \phi(m) - \lambda x \tt m ), 
$$
and 
\bb 
\Delta_t^1  & =(\dd \phi(\tm_{t+1}) - \lambda x_{t+1}) 
\tt 
( \tm_{t+1}  - \dd\phi^*(\lambda x_{t+1})), \\ 
\Delta^2_t & = 
 (\dd\phi(\tilde m_{t+1}) - \dd\phi(m_{t+1}))\tt  ( \tm_{t+1} - m_{t+1}), 
\ee 
where $\dd\phi^*$ is a subgradient of $\phi^*$. 
Then we have $\Delta_t^1\geq0$ and $\Delta_t^2 \geq 0$ from Lemma~\ref{lem:diff}, and 
\bb 
H(x_{t+1}, m_{t+1}) - H(x_t, m_t)
\leq 
- \epsilon (a  \Delta_t^1 
+ b \Delta_t^2)  + \frac{L\epsilon^2}{2} \norm{\dd \phi(\tm_{t+1}) - \lambda x_{t+1}}_2^2, 
\ee 
where 
\bb 
a = \frac{\lr \lambda \bone  }{ \lr \lambda (1-\bone) + (1-\btwo) } + 1 \geq 0, &&  
b = \frac{\bone(1-\btwo)}{(\btwo-\bone)(\lr \lambda (1-\bone) + (1-\btwo)) }\geq 0. 
\ee 
Hence, a telescoping sum yields 
$$
\frac{1}{T}\sum_{t=0}^{T-1} a \Delta_t^1 + b \Delta_t^2 
\leq \frac{H(x_0,m_0) - H(x_{T}, m_T)}{\epsilon T} + \frac{L\epsilon }{2} B_t, 
$$
where $B_t = \frac{1}{T}\sum_{t=0}^{T-1} \norm{\dd\phi(\tm_{t+1}) - \lambda x_{t+1}}^2_2$. 
\end{thm} 
Note that we used an implicit scheme in the update of $x_t$ in \eqref{equ:finiteupdate}. It is equivalent the explicit scheme with an adjusted learning rate:
\bb 
x_{t+1} & = x_t + \frac{\lr}{1+\lr \lambda }
(\dd \phi(\tilde m_{t+1}) - \lambda x_{t}). 
\ee 
\begin{proof} 
We follow the proof in the continuous-time case to find out a Lyapunov function for the discrete time update in \eqref{equ:finiteupdate}. 
We start with constructing the basic inequalities and work out the Lyapunov function backwardly. 
From Lemma~\ref{lem:key00}, we have 
\bbb  
(\dd \phi(\tm_{t+1}) - \lambda x_{t+1}) 
\tt 
( \dd\phi^*(\lambda x_{t+1}) - \tm_{t+1}) \leq 0. \label{equ:dphitm1}
\eee 
\bbb
& (\dd\phi(\tilde m_{t+1}) - \dd\phi(m_{t+1}))\tt  (m_{t+1} - \tm_{t+1})  \leq 0. 
\label{equ:dphitmtm2}
\eee 
Taking $a \times Eq.\eqref{equ:dphitm1}+b\times Eq\eqref{equ:dphitmtm2}$ for $a,b \geq 0$, we have 

\bb 
(\dd \phi(\tm_{t+1}) - \lambda x_{t+1})\tt 
(a ( \dd\phi^*(\lambda x_{t+1}) - \tm_{t+1}) &  + b  (m_{t+1} - \tm_{t+1}) )   + \cdots  \\ 
& + 
b (\dd\phi(m_{t+1}) - \lambda x_{t+1})\tt (-m_{t+1} + \tm_{t+1})
\leq 0.
\ee 
Plugging \eqref{equ:finiteupdate} yields  
\bb 
& (\dd \phi(\tm_{t+1}) -          \lambda x_{t+1})\tt 
(a  \dd\phi^*(\lambda x_{t+1})   
-((a+b) \bone -  b \btwo) m_t 
+ (a - (a+b) \bone  + b \btwo)\dd f(x_t) 
) \\ 
& - 
b (\btwo - \bone)  (\dd\phi(m_{t+1}) - \lambda x_{t+1})\tt 
(m_t + \dd f(x_t))
\leq 0 
\ee 
Define 
\bb 
H(x,m) = 
(a - c) f(x)  + 
\frac{a}{\lambda }  \phi^*(\lambda x)  + 
\frac{c}{\lambda } \phi(m)  - 
c  x \tt   m, &&\text{with}&& 
c = (a+b) \bone - b  \btwo, 
\ee 
and 
\bb 
 \hat \dd_x H_t = (a-c) \dd f(x_t) + a  \dd\phi^*(\lambda x_{t+1}) - c m_t, 
&&  \hat \dd_m H_t =  \frac{c}{\lambda }\dd \phi(m_{t+1}) - c x_{t+1}. 
\ee 
Then 
the inequality can be written into 
\bb 
\hat \dd_x H_t \tt 
\left ( \dd \phi(\tilde m_{t+1}) - \lambda x_{t+1} \right ) +  
\hat \dd_m H_t \tt \left ( \frac{b(\btwo - \bone)\lambda }{c} (- m_t - \dd f(x_t)) \right ) \leq 0. 
\ee 
Plugging the update rule of $x_{t+1}= x_t + \lr( \dd \phi(\tilde m_{t+1}) - \lambda x_{t+1}  )$ 
and $m_{t+1} - m_t = - (1-\btwo) (m_t + \dd f(x_t))$, we get 
\bb 
\hat \dd_x H_t \tt  
\left ( \frac{x_{t+1} - x_t}{\lr} \right )  +  
\hat \dd_m H_t \tt \left ( \frac{b(\btwo - \bone)\lambda }{c (1-\btwo)} (m_{t+1} - m_t) \right) \leq 0. 
\ee 
To make this coincide with the linear approximation of the difference $H(x_{t+1}, m_{t+1}) - H(x_{t}, m_t)$ (see Lemma~\ref{lem:diff}), 
we want 
$$
 \frac{b(\btwo - \bone)\lambda }{c(1-\btwo)} = \frac{1}{\lr}. 
$$
On the other hand, to make the coefficient of $f(x)$ in $H(x,m)$ equal to one, we want $a - c=1$. 
This yields the following equations on $a,b,c$:
\bb 
 c = (a+b) \bone - b \btwo , 
 && \frac{b(\btwo - \bone)\lambda }{c(1-\btwo)} = \frac{1}{\lr} ,  
 &&  a - c = 1, && a,b \geq 0.
\ee 
To solve this, let $c = z \epsilon (\btwo - \bone ) \lambda $ 
and $b = z (1-\btwo)$ for some $z \geq 0$ and plug them together with $a = c + 1$ into the first equations:
\bb 
 z \epsilon (\btwo - \bone ) \lambda = 
 ( z \epsilon (\btwo - \bone )\lambda  + 1 +z (1-\btwo) ) \bone - z (1-\btwo) \btwo. 
\ee 
We get 
\bb 
z 
& = \frac{\bone}{\epsilon (\btwo - \bone ) \lambda- 
  \epsilon (\btwo - \bone )\lambda \bone -  (1-\btwo)  \bone  +  (1-\btwo) \btwo} \\ 
  & = \frac{\bone}{\lr \lambda (\btwo-\bone)(1-\bone) + (1-\btwo)(\btwo - \bone)} \\ 
  & = \frac{\bone}{(\btwo-\bone)(\lr \lambda (1-\bone) + (1-\btwo)) } \geq 0.   
\ee 
Hence 
\bb 
b = \frac{\bone(1-\btwo)}{(\btwo-\bone)(\lr \lambda (1-\bone) + (1-\btwo)) }\geq0, && 
c = \frac{\lr \lambda \bone  }{ \lr \lambda (1-\bone) + (1-\btwo) }\geq0,  && a = c+1 \geq 0.
\ee 
In this case, we have 
\bb 
H(x,m) 
& = f(x) + \frac{1}{\lambda } \phi^*(\lambda x) + c (\phi^*(\lambda x) + \phi(m) - \lambda x \tt m ) \\ 
& = f(x) + \frac{1}{\lambda } \phi^*(\lambda x) + \frac{\lr \lambda \bone  }{ \lr \lambda (1-\bone) + (1-\btwo) }(\phi^*(\lambda x) + \phi(m) - \lambda x \tt m ), 
\ee 
and 
$$
\hat \dd_x H_t \tt  
\left ( \frac{x_{t+1} - x_t}{\lr} \right )  +  
\hat \dd_m H_t \tt \left (\frac{m_{t+1} - m_t}{\lr } \right) = - a  \Delta_t^1 
- b  \Delta_t^2  
\leq 0. 
$$
From Lemma~\ref{lem:diff}, we get 
\bb 
H(x_{t+1}, m_{t+1}) - H(x_t, m_t)
\leq 
-\epsilon ( a  \Delta_t^1 
+ b  \Delta_t^2  )  + \frac{L}{2} \norm{x_{t+1}- x_t}_2^2. 
\ee 
\end{proof}

\begin{lem}\label{lem:diff}
Let $H(x, m) = f(x) + \phi_1(x)  + \phi_2(m) -  \lambda x m $, where $f$ is $L$-smooth, and $\phi_1,\phi_2$ are  convex functions with subgradient 
 $\dd\phi_1$ and $\dd \phi_2$. 
Then 
$$
H(x_{t+1}, m_{t+1}) 
- H(x_{t}, m_t) 
\leq 
\hat \dd_x H_t \tt (x_{t+1}- x_t) 
+ \hat \dd_m H_t \tt (m_{t+1} - m_t) 
+ \frac{L}{2} \norm{x_{t+1} - x_t}_2^2,
$$
where 
\bb 
& \hat \dd_x H_t = \dd f(x_t) + \dd \phi_1(x_{t+1})- \lambda m_t \\ 
& \hat \dd_m H_t
= \dd \phi_2(m_{t+1}) - \lambda x_{t+1}. 
\ee 
Note the use of $x_t$ vs. $x_{t+1}$ and $m_t$ vs. $m_{t+1}$ in $\hat \dd_x H_t$ and $\hat \dd_m H_t$. 
\end{lem} 
\begin{proof}
We have 
\bb 
f(x_{t+1}) - f(x_t) & \leq \dd f(x_t)\tt (x_{t+1} - x_t) + \frac{L}{2} \norm{x_{t+1} - x_t}^2_2 \\ 
\phi_1(x_{t+1}) - \phi_1(x_t) &\leq \dd\phi_1(x_{t+1}) \tt (x_{t+1} - x_t) \\
\phi_2(m_{t+1}) - \phi_2(m_t) &\leq \dd\phi_2(m_{t+1}) \tt (m_{t+1} - m_t) \\
x_{t+1}\tt m_{t+1} - x_t \tt m_t 
& = m_t \tt (x_{t+1} - x_t) 
+ x_{t+1}\tt (m_{t+1} - m_t). 
\ee 
Summing them together yields the result. 
\end{proof}


\begin{thm}
Under the same conditions of Theorem~\ref{thm:discretemain}, 
for any two integers $s \leq t,$ 
$$
\dist(\lambda x_t, \dom \phi^*) \leq \left (\frac{1}{1+\epsilon \lambda }\right)^{t-s} \dist(\lambda x_s, \dom \phi^*), ~~~\forall s\leq t.
$$
\end{thm}
\begin{proof} 
Rewriting the update into the explicit form: 
$$
x_{t+1} = \frac{1}{1+\epsilon \lambda } x_t +  \frac{\epsilon}{1+\epsilon\lambda} \dd\phi(\tilde m_{t+1}). 
$$
Unrolling this update  yields,
with $w_{s\to t} = \left ( \frac{1}{1+\epsilon \lambda}\right )^{t-s},$ 
\bb 
\lambda x_{t}
& =( 1- w_{s\to t}) z_{s\to t}
+ w_{s\to t} x_s, ~~~~~~ 
~~~~ 
z_{s\to t} = \frac{\sum_{k=s+1}^{t} w_{k\to t} \dd\phi(\tilde m_k)}{\sum_{k=s+1}^t w_{k\to t}}. 
\ee
We have $\dd\phi(\tilde m_k) \in \dom \phi^*$ from Lemma~\ref{lem:dom} and $\dom \phi^*$ is convex. Hence $z_{s\to t} $, as the convex combination of $\{\dd\phi(\tilde m_k\}_k$, belongs to $\dom \phi^*$. 
For any $\eta >0$, let $\lambda \hat x_s \in \dom \phi^*$ to the point satisfying $\norm{\lambda \hat x_s - \lambda x_s} \leq \dist(\lambda x_s, \dom \phi^*) + \eta$. 
Hence, 
\bb 
\mathrm{dist}(\lambda x_t, ~ \dom \phi^*) 
& = \inf_{z \in \dom \phi^*} \norm{\lambda x_t  - z }  \\
& \leq \norm{\lambda x_t - (1-w_{s\to t}) z_{s\to t} + w_{s\to t} \lambda \hat x_s)} \\
& =  w_{s\to t} \norm{\lambda x_s - \lambda \hat x_s} \\
& \leq\left ( \frac{1}{1+\epsilon \lambda}\right )^{s-t}  (\dist(\lambda x_s, \dom \phi^*) + \eta). 
\ee 
Taking $\eta \to 0$ yields the result. 
 

\end{proof}

\subsection{Analysis with Stochastic Gradient for Lion-$\phi$}

In this section, we are going to have the convergence analysis of discrete time Lion-$\phi$. The proof idea is adapted for section~\ref{sec:discreteapp}, by defining the same Hamiltonian function, we obtain the bound for $\Delta_t^1$ and $\Delta_t^2$. 

Compared with the deterministic case, 
the main challenge 
is to bound an additional correlation term 
due to 
the stochastic gradient at each iteration $t$:  
\begin{align} \label{equ:covgm} 
V_t \defeq \cov(g_t, ~ \dd\phi(\tilde m_{t+1}))
= \cov(g_t, ~ \dd\phi(\beta_1 m_t + (1-\beta_1) g_t)),
\end{align} 
where $\cov(X,Y) = \E[(X-\E[X])\tt (Y-\E[Y])]$.

\begin{mydef}
For a random variable $X$ on $\RR^d$, its (trace of) variance $\var(X)$, when exists, is defined as  
\bb 
\var(X) = \E[\norm{X - \E[X]}^2_2]
\ee
\end{mydef} 

\begin{ass}\label{ass:batch}
Assume 
\bb 
\var(g_t) \leq \frac{v_{\max}}{n_{batch}},
 \ee 
 where $n_{batch}$ represents the batch size. 
\end{ass} 

\begin{ass}\label{ass::sample}
    $\mathcal{D}$ is the data distribution, the stochastic sample $\xi_t \sim \mathcal{D}$ is i.i.d., given a function $f(x;\xi)$, the gradient $\dd f(x;\xi)$ is taken with respect to variable $x$, and $\E[\dd f(x, \xi)] = \dd f(x)$
\end{ass}

\begin{thm}\label{thm:sgd}
    Under the assumptions delineated in \ref{ass::sample} and \ref{ass:batch}, consider a function $f\colon \mathbb{R}^d\to \mathbb{R}$ that is $L$-smooth. Additionally, let $\phi\colon \mathbb{R}^d\to \mathbb{R}$ be a closed and convex function, consider the following scheme: 
\bbb \label{equ:finiteupdate}
\begin{split}
m_{t+1} & = \btwo m_t  - (1-\btwo) g_t \\ 
\tilde m_{t+1} & = \bone m_{t} - (1-\bone) g_t \\ 
x_{t+1} & = x_t + \lr
(\dd \phi(\tilde m_{t+1}) - \lambda x_{t+1}), 
\end{split}
\eee
where $g_t = \dd f(x_t;\xi_t)$ as shown in ~\ref{ass::sample}, $m_0, g_1, \ldots, g_t, \ldots$ are random variables with $\E[g_t] = \dd f(x_t)$. 
$\dd\phi$ is a weak gradient of $\phi$ with $\dd \phi(0) = 0$, $\left\|\dd \phi(x) - \dd \phi(y)\right\| \leq L_\phi \left\|x-y\right\|, \forall x,y \in \RR^d$, and
$\bone,\btwo \in (0,1)$, and $\btwo>\bone$, and $\epsilon, \lambda > 0$. 

Let $\phi^*$ be the conjugate function of $\phi$. 
Define the following Lyapunov function:
$$
H(x, m ) = f(x) + \frac{1}{\lambda } \phi^*(\lambda x) + \frac{\epsilon\bone}{\epsilon \lambda  (1-\bone) + (1-\btwo)}(\phi^*(\lambda x) + \phi(m) - \lambda x \tt m ), 
$$
and
\bb
\Delta_t^1  & =(\dd \phi(\tm_{t+1}) - \lambda x_{t+1})
\tt 
( \tm_{t+1}  - \dd\phi^*(\lambda x_{t+1})), \\ 
\Delta^2_t & = 
 (\dd\phi(\tilde m_{t+1}) - \dd\phi(m_{t+1}))\tt  ( \tm_{t+1} - m_{t+1}), 
\ee 
where $\dd\phi^*$ is a subgradient of $\phi^*$. 
Then we have $\Delta_t^1\geq0$ and $\Delta_t^2 \geq 0$ from Lemma~\ref{lem:diff}, and 
\bb 
\E\left[H(x_{t+1}, m_{t+1}) - H(x_t, m_t)\right]
\leq 
\E\left[- \epsilon (a  \Delta_t^1 
+ b \Delta_t^2)  + \frac{L\epsilon^2}{2} \norm{\dd \phi(\tm_{t+1}) - \lambda x_{t+1}}_2^2\right] \\
+ \epsilon \frac{L_{\phi}}{1+\lambda \epsilon} (1-\beta_1) \frac{v_{max}}{n_{batch}} + \frac{L_{\phi}}{1+\lambda \epsilon} \sqrt{\frac{(1-\btwo)}{(1+\btwo)}}\frac{v_{max}}{n_{batch}} 
\ee 
where 
\bb 
a = \frac{\lr \lambda \bone }{\lr \lambda (1-\bone) + (1-\btwo) } + 1 \geq 0, &&  
b = \frac{\bone(1-\btwo)}{(\btwo-\bone)(\lr \lambda (1-\bone) + (1-\btwo)) }\geq 0. 
\ee 
$v_{\max}$, $n_{batch}$ are defined in ~\ref{ass:batch}

Hence, a telescoping sum yields 
$$
\frac{1}{T}\sum_{t=0}^{T-1} \E\left[a \Delta_t^1 + b \Delta_t^2 \right]
\leq \E\left[\frac{H(x_0,m_0) - H(x_{T}, m_T)}{\epsilon T} + \frac{L\epsilon }{2} B_t + \frac{C_t}{n_{\text{batch}}}\right], 
$$
where $B_t = \frac{1}{T}\sum_{t=1}^T \norm{\dd\phi(\tm_{t+1}) - \lambda x_{t+1}}^2_2$, and 
$C_t = \left(\frac{L_{\phi}}{1+\lambda \epsilon} (1-\beta_1) + \frac{L_{\phi}}{1+\lambda \epsilon} \sqrt{\frac{(1-\btwo)}{(1+\btwo)}}\right) v_{max} $.
\end{thm}

\begin{proof}
    The proof is a simple extended variant of ~\ref{thm:discretemain}.
Following the proof of Theorem~\ref{thm:discreteapp}, 
define 
\bb 
H(x,m) = 
(a - c) f(x)  + 
\frac{a}{\lambda }  \phi^*(\lambda x)  + 
\frac{c}{\lambda } \phi(m)  - 
c  x \tt   m, &&\text{with}&& 
c = (a+b) \bone - b  \btwo,
\ee
where 
\bb 
a = \frac{\lr \lambda \bone }{\lr \lambda (1-\bone) + (1-\btwo) } + 1 \geq 0, &&  
b = \frac{\bone(1-\btwo)}{(\btwo-\bone)(\lr \lambda (1-\bone) + (1-\btwo)) }\geq 0, && c = a - 1. 
\ee 
By the definition of $\Delta_t^1, \Delta_t^2$, we have
\bbb \label{eqn:delta}
&a \Delta_t^1 + b \Delta_t^2 \nonumber\\
&= a  (\dd \phi(\tm_{t+1}) - \lambda x_{t+1})\tt (\tm_{t+1} - \dd\phi^*(\lambda x_{t+1})) \nonumber\\
&\quad \quad \quad \quad + b  (\dd\phi(\tilde m_{t+1}) - \dd\phi(m_{t+1}))\tt  ( \tm_{t+1} - m_{t+1}) \nonumber\\
&=(\dd \phi(\tm_{t+1}) - \lambda x_{t+1})\tt 
(a ( \dd\phi^*(\lambda x_{t+1}) - \tm_{t+1})  + b  ( \tm_{t+1}-m_{t+1}) ) \nonumber \\ 
& \quad \quad \quad \quad  + b (\dd\phi(m_{t+1}) - \lambda x_{t+1})\tt (m_{t+1} - \tm_{t+1})\nonumber \\
& =-(\dd \phi(\tm_{t+1}) - \lambda x_{t+1})\tt (a  \dd\phi^*(\lambda x_{t+1})   
-((a+b) \bone -  b \btwo) m_t + (a - (a+b) \bone  + b \btwo)\dd f(x_t) ) \nonumber\\ 
& \quad \quad \quad \quad - b \frac{\beta_2 - \beta_1}{1 -\beta_2} \frac{\lambda}{c} (\frac{c}{\lambda}\dd\phi(m_{t+1}) - c x_{t+1})\tt 
(m_{t+1} - m_t) \nonumber\\
&=-\left[(a-c) g_t + a \dd \phi^*(\lambda x_{t+1}) - c m_t\right]\tt (\dd \phi(\tm_{t+1}) - \lambda x_{t+1}) \nonumber\\
    &\quad \quad \quad \quad - \frac{1}{\epsilon}\left[\frac{c}{\lambda} \dd \phi(m_{t+1}) - c x_{t+1}\right]\tt (m_{t+1} - m_t)\nonumber\\
&=-\frac{1}{\epsilon}\left[(a-c) g_t + a \dd \phi^*(\lambda x_{t+1}) - c m_t\right]\tt (x_{t+1} - x_t)\nonumber \\
    &\quad\quad \quad \quad - \frac{1}{\epsilon}\left[\frac{c}{\lambda} \dd \phi(m_{t+1}) - c x_{t+1}\right]\tt (m_{t+1} - m_t)
\eee

By Lemma~\ref{lem:diff}, 
$$
H(x_{t+1}, m_{t+1}) 
- H(x_{t}, m_t) 
\leq 
\hat \dd_x H_t \tt (x_{t+1}- x_t) 
+ \hat \dd_m H_t \tt (m_{t+1} - m_t) 
+ \frac{L}{2} \norm{x_{t+1} - x_t}_2^2,
$$
where 
\bb 
& \hat \dd_x H_t = (a-c) \dd f(x_t) + a  \dd\phi^*(\lambda x_{t+1}) - c m_t,  \\ 
&  \hat \dd_m H_t 
=  \frac{c}{\lambda }\dd \phi(m_{t+1}) - c x_{t+1} = \frac{c}{\epsilon\lambda} (\hat V_{x,t} 
 - \dd\phi(\tilde m_{t+1}) + \dd\phi(m_{t+1} ))
\ee 
with
\bb 
& V_{x,t} =x_{t+1} - x_t    = \epsilon (\dd \phi(\tilde m_{t+1}) - \lambda x_{t+1}) \\ 
& V_{m,t}  = m_{t+1} - m_t   = - (1-\beta_2) (g_t - m_t) \\
&\tilde m_{t+1} - m_{t+1} =  - (\beta_2 - \beta_1) (g_t - m_t) = - (\beta_2 - \beta_1) V_{m,t} \\ 
& \hat V_{m,t} = 
 - \dd\phi(\tilde m_{t+1} ) + \dd\phi(m_{t+1})  \\
\ee
This gives 
\bb 
& H(x_{t+1}, m_{t+1}) 
- H(x_{t}, m_t)  \\ 
& \leq 
\hat \dd_x H_t \tt (x_{t+1}- x_t) 
+ \hat \dd_m H_t \tt (m_{t+1} - m_t) 
+ \frac{L}{2} \norm{x_{t+1} - x_t}_2^2 \\
& 
\ee

Hence,
    \bb
    H(x_{t+1}, m_{t+1}) - H(x_t, m_t) &\leq \left[(a-c) \dd f(x_t) + a \dd \phi^*(\lambda x_{t+1}) - c m_t\right]\tt (x_{t+1} - x_t) \\
    &\quad + \left[\frac{c}{\lambda} \dd \phi(m_{t+1}) - c x_{t+1}\right]\tt (m_{t+1} - m_t) + \frac{L}{2}\|x_{t+1} - x_t\|_2^2 \\
    &= \left[(a-c) g_t + a \dd \phi^*(\lambda x_{t+1}) - c m_t\right]\tt (x_{t+1} - x_t) \\
    &\quad + \left[\frac{c}{\lambda} \dd \phi(m_{t+1}) - c x_{t+1}\right]\tt (m_{t+1} - m_t) + \frac{L}{2}\|x_{t+1} - x_t\|_2^2\\
    &\quad + \epsilon (a-c)(\dd f(x_t) - g_t)\tt (\dd \phi(\tm_{t+1}) - \lambda x_{t+1}) \\
    &=- \epsilon (a  \Delta_t^1+ b \Delta_t^2) + \frac{L}{2}\|x_{t+1} - x_t\|_2^2 \ant{by equation~\ref{eqn:delta}}\\
    &\quad + \epsilon (a-c)(\dd f(x_t) - g_t)\tt (\dd \phi(\tm_{t+1}) - \lambda x_{t+1})
    \ee 
    It suffices to bound $\E \left[(\dd f(x_t) - g_t)\tt (\dd \phi(\tm_{t+1}) - \lambda x_{t+1})\right]$. 
    
    Note that 
    \bb
    &\E \left[(\dd f(x_t) - g_t)\tt (\dd \phi(\tm_{t+1}) -\lambda x_{t+1})\right] \\
    &= \E \left[(\dd f(x_t) - g_t)\tt (\frac{1}{1+\lambda\epsilon}\dd \phi(\tm_{t+1}) - \frac{\lambda}{1+\lambda \epsilon} x_{t})\right] \\
    &=\frac{1}{1+\lambda\epsilon}\E \left[(\dd f(x_t) - g_t)\tt \dd \phi(\tm_{t+1})\right] + \frac{\lambda}{1+\lambda \epsilon}\E \left[(\dd f(x_t) - g_t)\tt  x_{t}\right] 
    \ee
    By Assumption~\ref{ass::sample}, 
    \bb
    \E\left[ (\dd f(x_t) - g_t)\tt \lambda x_{t}) \right] &= \lambda\E_{x_t} \left[\E_{\xi_t}\left[ (\dd f(x_t) - \dd f(x_t, \xi_t))\tt x_t ~|~ x_{t} \right]\right]\\
    & = 0\ant{by \ref{ass::sample}~ $\E[\dd f(x, \xi)] = \dd f(x)$}
    \ee
    
    Next, let us bound $\E\left[(\dd f(x_t) - g_t)\tt \dd \phi(\tm_{t+1})\right]$. 
\bb
\E \left[\left(\dd f(x_t) - g_t\right)\tt \dd \phi(\tm_{t+1})\right] & =  \E \left[\left(\dd f(x_t) - g_t \right)\tt \dd \phi(\beta_1 m_{t} - (1-\beta_1) g_t)\right]\\
& \leq  L_\phi (1-\beta_1) \var(g_t) + L_\phi \sqrt{\var(\bone m_t)\cdot \var(g_t)} \ant{by \ref{lem:XY}}\\
& \leq  L_\phi (1-\beta_1) \frac{v_{max}}{n_{batch}} + L_\phi \sqrt{\frac{(1-\btwo)}{(1+\btwo)}}\frac{v_{max}}{n_{batch}} \ant{by \ref{lem:XY}}
\ee
Hence,
 \bb
    &\E \left[(\dd f(x_t) - g_t)\tt (\dd \phi(\tm_{t+1}) -\lambda x_{t+1})\right] \\
    &=\frac{1}{1+\lambda\epsilon}\E \left[(\dd f(x_t) - g_t)\tt \dd \phi(\tm_{t+1})\right] + \frac{\lambda}{1+\lambda \epsilon}\E \left[(\dd f(x_t) - g_t)\tt  x_{t+1}\right] \\
    &\leq \frac{L_{\phi}}{1+\lambda \epsilon} (1-\beta_1) \frac{v_{max}}{n_{batch}} + \frac{L_{\phi}}{1+\lambda \epsilon} \sqrt{\frac{(1-\btwo)}{(1+\btwo)}}\frac{v_{max}}{n_{batch}} 
\ee
\end{proof}

\begin{lem}\label{lem:XY}
Let $X,Y$ be two $\RR^d$-valued random variables with $\var(X)<+\infty$ and $\var(Y)<+\infty$,  and assume $\phi$ yields a weak derivative $\dd\phi$. We have
$$\E[(Y - \E Y)\tt \dd\phi(X+\epsilon Y)] \leq L_\phi \epsilon \var(Y) + L_\phi \sqrt{\var(X)\cdot \var(Y)}
$$
\end{lem}
\begin{proof}
\bb 
&\E[ (Y-\E [Y])\tt \dd\phi(X + \eps Y) ]\\
& = \E [(Y-\E [Y]) \tt \left(\dd \phi(X + \epsilon Y) - \dd \phi(\E[X] + \epsilon \E[Y] \right) ] \\
& \leq \sqrt{\E \left\|Y-\E [Y]\right\|^2} \sqrt{\E\left\|\dd \phi(X + \epsilon Y) - \dd \phi(\E[X] + \epsilon \E[Y]) \right\|^2} \\
& \leq \sqrt{\E \left\|Y-\E [Y]\right\|^2} \sqrt{L_{\phi}^2\E\left\|X + \epsilon Y - \E[X] - \epsilon \E[Y] \right\|^2} \\
& \leq L_\phi \sqrt{\E \left\|Y-\E [Y]\right\|^2} \left(\sqrt{\E\left\|X - \E[X]\right\|^2} + \sqrt{\epsilon^2 \E\left\|Y - \E[Y] \right\|^2}\right)\\
& = L_\phi \epsilon \E\left\|Y-\E [Y]\right\|^2 + L_\phi \sqrt{\E \left\|Y-\E [Y]\right\|^2}\sqrt{\E\left\|X - \E[X]\right\|^2}\\
& = L_\phi \epsilon \var(Y) + L_\phi \sqrt{\var(X)\cdot \var(Y)}
\ee 
\end{proof}

\begin{lem}[Cumulative error of stochastic gradient~\cite{bernstein_signsgd_2018}]\label{lem:martingale}
Following the same setting in theorem~\ref{thm:sgd}, denote $\delta_l = g_l - \dd f(x_l)$, for any $k<\infty$ and fixed weight $-\infty < \alpha_1,...,\alpha_k< \infty$, $\sum_{l=1}^k \alpha_l \delta_l$ is a Martingale. In particular,
	$$
	\E\left[\left[\sum_{l=1}^k  \alpha_l  \delta_l\right]^2\right] \leq  \sum_{l=1}^k \alpha_l^2 {  \sigma}^2.
	$$
\end{lem}
\begin{proof}
	We simply check the definition of a Martingale. Denote $Y_k:= \sum_{l=1}^k\alpha_l \delta_l$. 
	First, we have that
	\begin{align*}
	\E[|Y_k|] &= \E\left[\abs{\sum_{l=1}^k\alpha_l \delta_l}\right] \\  
    &\leq \sum_l  |\alpha_l|\E[|\delta_l|] &\text{triangle inequality}\\
    &= \sum_l  |\alpha_l|  \E[\E[|\delta_l| | x_l]] &\text{law of total probability}\\
   &\leq \sum_l  |\alpha_l|\E[\sqrt{\E[ \delta_l^2  |x_l]} ]&\text{Jensen's inequality}\\
   &\leq \sum_l |\alpha_l|   \sigma  <\infty 
	\end{align*}
	Second, again using the law of total probability,
	\begin{align*}
	\E[Y_{k+1} | Y_1,...,Y_k]  &= \E\left[\sum_{l=1}^{k+1}\alpha_l \delta_l  \middle|  \alpha_1 \delta_1, ...,  \alpha_k \delta_k \right]  \\
	&=  Y_k  +  \alpha_{k+1}\E\left[ \delta_{k+1}  \middle|  \alpha_1 \delta_1, ...,  \alpha_k \delta_k  \right]\\
	&= Y_k  +   \alpha_{k+1}\E\left[  \E\left[ \delta_{k+1}  \middle|  x_{k+1},  \alpha_1 \delta_1, ...,  \alpha_k \delta_k  \right] | \alpha_1 \delta_1, ...,  \alpha_k \delta_k \right]\\
	&= Y_k  + \alpha_{k+1}\E\left[  \E\left[ \delta_{k+1}  \middle|  x_{k+1}\right] | \alpha_1 \delta_1, ...,  \alpha_k \delta_k \right]\\
	&=Y_k
	\end{align*}
	This completes the proof that it is indeed a Martingale. We now make use of the properties of Martingale difference sequences to establish a variance bound on the Martingale.
	\begin{align*}
	\E[[\sum_{l=1}^k \alpha_l \delta_l ]^2 ]  &=  \sum_{l=1}^k \E[  \alpha_l^2 \delta_l^2]   +  2\sum_{l<j} \E[\alpha_l\alpha_j \delta_l \delta_j]\\
	&=\sum_{l=1}^k \alpha_l^2  \E[\E[\delta_l^2| \delta_1,...,\delta_{l-1} ]]   +   2\sum_{l<j} \alpha_l\alpha_j  \E\Big[\delta_l \E\big[ \E[\delta_j | \delta_1,...,\delta_{j-1}] \big| \delta_l \big]\Big]\\
    &=\sum_{l=1}^k \alpha_l^2  \E[\E[\E[\delta_l^2| x_l,\delta_1,...,\delta_{l-1} ]| \delta_1,...,\delta_{l-1} ]]   +   0\\
	&=\sum_{l=1}^k \alpha_l^2   \sigma^2.
	\end{align*}
\end{proof}

The consequence of this lemma is that we are able to treat $\delta_1,...,\delta_k$ as if they are  independent, even though 
they are not---clearly $\delta_l$ is dependent on $\delta_1,...,\delta_{l-1}$ through $x_l$. By Lemma~\ref{lem:martingale}, we can compute the variance of momentum $m_t$, 
\bb
\var(m_t) &= (1-\btwo)^2\E \left\|\sum_{i=1}^t \btwo^{t-i} \delta_i\right\|^2 \\
&\leq (1-\btwo)^2\E \sum_{i=1}^t \btwo^{2t-2i} \left\|\delta_i\right\|^2\\
&= \frac{(1-\btwo)v_{max}}{(1+\btwo)n_{batch}}
\ee

\subsection{Analysis with Stochastic Gradient LION}

\begin{thm}\label{pix_thm:sgd}
    Under the assumptions delineated in \ref{ass::sample} and \ref{ass:batch}, consider a function $f\colon \mathbb{R}^d\to \mathbb{R}$ that is $L$-smooth. Consider the following scheme: 
\bbb \label{equ:finiteupdate}
\begin{split}
m_{t+1} & = \btwo m_t  - (1-\btwo) g_t \\ 
\tilde m_{t+1} & = \bone m_{t} - (1-\bone) g_t \\ 
x_{t+1} & = x_t + \lr
(sign(\tilde m_{t+1}) - \lambda x_{t+1}), 
\end{split}
\eee
where $g_t = \dd f(x_t;\xi_t)$ as shown in ~\ref{ass::sample}, $m_0, g_1, \ldots, g_t, \ldots$ are random variables with $\E[g_t] = \dd f(x_t)$. $\bone,\btwo \in (0,1)$, and $\btwo>\bone$, and $\epsilon, \lambda > 0$. 
 
Define the following Lyapunov function:
$$
H(x, m ) = f(x) + \frac{1}{\lambda } \|\lambda x\|^* + \frac{\bone}{\epsilon \lambda  (1-\bone) + (1-\btwo)}(\|\lambda x\|^* + \|m\| - \lambda x \tt m ), 
$$
and
\bb
\Delta_t^1  & =(sign(\tm_{t+1}) - \lambda x_{t+1})
\tt 
( \tm_{t+1}  - sign^*(\lambda x_{t+1})), \\ 
\Delta^2_t & = 
 (sign(\tilde m_{t+1}) - sign(m_{t+1}))\tt  ( \tm_{t+1} - m_{t+1}), 
\ee 
where $sign^*$ is a subgradient of $\phi^*$. 
Then we have $\Delta_t^1\geq0$ and $\Delta_t^2 \geq 0$ from Lemma~\ref{lem:diff}, and 
\bb 
\E\left[H(x_{t+1}, m_{t+1}) - H(x_t, m_t)\right]
\leq 
\E\left[- \epsilon (a  \Delta_t^1 
+ b \Delta_t^2)  + \frac{L\epsilon^2}{2} \norm{sign(\tm_{t+1}) - \lambda x_{t+1}}_2^2\right] 
+ \epsilon \frac{1}{1+\lambda\epsilon}\frac{\sqrt{d \cdot v_{max}}}{\sqrt{n_{batch}}}
\ee 
where 
\bb 
a = \frac{\lr \lambda \bone }{\lr \lambda (1-\bone) + (1-\btwo) } + 1 \geq 0, &&  
b = \frac{\bone(1-\btwo)}{(\btwo-\bone)(\lr \lambda (1-\bone) + (1-\btwo)) }\geq 0. 
\ee 
$v_{max}$, $n_{batch}$ are defined in ~\ref{ass:batch}

Hence, a telescoping sum yields 
$$
\frac{1}{T}\sum_{t=0}^{T-1} \E\left[a \Delta_t^1 + b \Delta_t^2 \right]
\leq \E\left[\frac{H(x_0,m_0) - H(x_{T}, m_T)}{\epsilon T} + \frac{L\epsilon }{2} B_t + \frac{1}{1+\lambda\epsilon}\frac{\sqrt{d \cdot v_{max}}}{\sqrt{n_{\text{batch}}}}\right], 
$$
where $B_t = \frac{1}{T}\sum_{t=1}^T \norm{sign(\tm_{t+1}) - \lambda x_{t+1}}^2_2$
\end{thm}

\begin{proof}
Define 
\bb 
H(x,m) = 
(a - c) f(x)  + 
\frac{a}{\lambda }  \|\lambda x\|^*  + 
\frac{c}{\lambda } \|m\|  - 
c  x \tt   m, &&\text{with}&& 
c = (a+b) \bone - b  \btwo,
\ee
where 
\bb 
a = \frac{\lr \lambda \bone }{\lr \lambda (1-\bone) + (1-\btwo) } + 1 \geq 0, &&  
b = \frac{\bone(1-\btwo)}{(\btwo-\bone)(\lr \lambda (1-\bone) + (1-\btwo)) }\geq 0, && c = a - 1. 
\ee 
By the definition of $\Delta_t^1, \Delta_t^2$, we have
\bbb \label{eqn:delta}
&a \Delta_t^1 + b \Delta_t^2 \nonumber\\
&= a  (sign(\tm_{t+1}) - \lambda x_{t+1})\tt (\tm_{t+1} - sign^*(\lambda x_{t+1})) \nonumber\\
&\quad \quad \quad \quad + b  (sign(\tilde m_{t+1}) - sign(m_{t+1}))\tt  ( \tm_{t+1} - m_{t+1}) \nonumber\\
&=(sign(\tm_{t+1}) - \lambda x_{t+1})\tt 
(a ( sign^*(\lambda x_{t+1}) - \tm_{t+1})  + b  ( \tm_{t+1}-m_{t+1}) ) \nonumber \\ 
& \quad \quad \quad \quad  + b (sign(m_{t+1}) - \lambda x_{t+1})\tt (m_{t+1} - \tm_{t+1})\nonumber \\
& =-(sign(\tm_{t+1}) - \lambda x_{t+1})\tt (a  sign^*(\lambda x_{t+1})   
-((a+b) \bone -  b \btwo) m_t + (a - (a+b) \bone  + b \btwo)\dd f(x_t) ) \nonumber\\ 
& \quad \quad \quad \quad - b \frac{\beta_2 - \beta_1}{1 -\beta_2} \frac{\lambda}{c} (\frac{c}{\lambda}sign(m_{t+1}) - c x_{t+1})\tt 
(m_{t+1} - m_t) \nonumber\\
&=-\left[(a-c) g_t + a sign^*(\lambda x_{t+1}) - c m_t\right]\tt (sign(\tm_{t+1}) - \lambda x_{t+1}) \nonumber\\
    &\quad \quad \quad \quad - \frac{1}{\epsilon}\left[\frac{c}{\lambda} sign(m_{t+1}) - c x_{t+1}\right]\tt (m_{t+1} - m_t)\nonumber\\
&=-\frac{1}{\epsilon}\left[(a-c) g_t + a sign^*(\lambda x_{t+1}) - c m_t\right]\tt (x_{t+1} - x_t)\nonumber \\
    &\quad\quad \quad \quad - \frac{1}{\epsilon}\left[\frac{c}{\lambda} sign(m_{t+1}) - c x_{t+1}\right]\tt (m_{t+1} - m_t)
\eee

By Lemma~\ref{lem:diff}, 
$$
H(x_{t+1}, m_{t+1}) 
- H(x_{t}, m_t) 
\leq 
\hat \dd_x H_t \tt (x_{t+1}- x_t) 
+ \hat \dd_m H_t \tt (m_{t+1} - m_t) 
+ \frac{L}{2} \norm{x_{t+1} - x_t}_2^2,
$$
where 
\bb 
& \hat \dd_x H_t = (a-c) \dd f(x_t) + a  sign^*(\lambda x_{t+1}) - c m_t,  \\ 
&  \hat \dd_m H_t 
=  \frac{c}{\lambda }sign(m_{t+1}) - c x_{t+1} = \frac{c}{\epsilon\lambda} (\hat V_{x,t} 
 - sign(\tilde m_{t+1}) + sign(m_{t+1} ))
\ee 
with
\bb 
& V_{x,t} =x_{t+1} - x_t    = \epsilon (sign(\tilde m_{t+1}) - \lambda x_{t+1}) \\ 
& V_{m,t}  = m_{t+1} - m_t   = - (1-\beta_2) (g_t - m_t) \\
&\tilde m_{t+1} - m_{t+1} =  - (\beta_2 - \beta_1) (g_t - m_t) = - (\beta_2 - \beta_1) V_{m,t} \\ 
& \hat V_{m,t} = 
 - sign(\tilde m_{t+1} ) + sign(m_{t+1})  \\
\ee
This gives 
\bb 
& H(x_{t+1}, m_{t+1}) 
- H(x_{t}, m_t)  \\ 
& \leq 
\hat \dd_x H_t \tt (x_{t+1}- x_t) 
+ \hat \dd_m H_t \tt (m_{t+1} - m_t) 
+ \frac{L}{2} \norm{x_{t+1} - x_t}_2^2 \\
& 
\ee

Hence,
    \bb
    H(x_{t+1}, m_{t+1}) - H(x_t, m_t) &\leq \left[(a-c) \dd f(x_t) + a sign^*(\lambda x_{t+1}) - c m_t\right]\tt (x_{t+1} - x_t) \\
    &\quad + \left[\frac{c}{\lambda} sign(m_{t+1}) - c x_{t+1}\right]\tt (m_{t+1} - m_t) + \frac{L}{2}\|x_{t+1} - x_t\|_2^2 \\
    &= \left[(a-c) g_t + a sign^*(\lambda x_{t+1}) - c m_t\right]\tt (x_{t+1} - x_t) \\
    &\quad + \left[\frac{c}{\lambda} sign(m_{t+1}) - c x_{t+1}\right]\tt (m_{t+1} - m_t) + \frac{L}{2}\|x_{t+1} - x_t\|_2^2\\
    &\quad + \epsilon (a-c)(\dd f(x_t) - g_t)\tt (sign(\tm_{t+1}) - \lambda x_{t+1}) \\
    &=- \epsilon (a  \Delta_t^1+ b \Delta_t^2) + \frac{L}{2}\|x_{t+1} - x_t\|_2^2 \ant{by equation~\ref{eqn:delta}}\\
    &\quad + \epsilon (a-c)(\dd f(x_t) - g_t)\tt (sign(\tm_{t+1}) - \lambda x_{t+1})
    \ee 
    It suffices to bound $\E \left[(\dd f(x_t) - g_t)\tt (sign(\tm_{t+1}) - \lambda x_{t+1})\right]$. 
    
    Note that 
    \bb
    &\E \left[(\dd f(x_t) - g_t)\tt (sign(\tm_{t+1}) -\lambda x_{t+1})\right] \\
    &= \E \left[(\dd f(x_t) - g_t)\tt (\frac{1}{1+\lambda\epsilon}sign(\tm_{t+1}) - \frac{\lambda}{1+\lambda \epsilon} x_{t})\right] \\
    &=\frac{1}{1+\lambda\epsilon}\E \left[(\dd f(x_t) - g_t)\tt sign(\tm_{t+1})\right] + \frac{\lambda}{1+\lambda \epsilon}\E \left[(\dd f(x_t) - g_t)\tt  x_{t}\right] 
    \ee
    By Assumption~\ref{ass::sample}, 
    \bb
    \E\left[ (\dd f(x_t) - g_t)\tt \lambda x_{t}) \right] &= \lambda\E_{x_t} \left[\E_{\xi_t}\left[ (\dd f(x_t) - \dd f(x_t, \xi_t))\tt x_t ~|~ x_{t} \right]\right]\\
    & = 0\ant{by \ref{ass::sample}~ $\E[\dd f(x, \xi)] = \dd f(x)$}
    \ee
    
    Next, we can use \ref{lem:XY} to bound $\E\left[(\dd f(x_t) - g_t)\tt sign(\tm_{t+1})\right]$. 

    \bb
    \E \left[\left(\dd f(x_t) - g_t\right)\tt sign(\tm_{t+1})\right] & =  \E \left[\left(\dd f(x_t) - g_t \right)\tt sign(\beta_1 m_{t} - (1-\beta_1) g_t)\right]\\
    & \leq  \sqrt{d \cdot \var(g_t)} \ant{by \ref{lem:XY} }\\
    & \leq \sqrt{\frac{d \cdot v_{max}}{n_{batch}}} \ant{by \ref{ass:batch} }
    \ee
Hence,
\bb
    &\E \left[(\dd f(x_t) - g_t)\tt (sign(\tm_{t+1}) -\lambda x_{t+1})\right] \\
    &=\frac{1}{1+\lambda\epsilon}\E \left[(\dd f(x_t) - g_t)\tt sign(\tm_{t+1})\right] + \frac{\lambda}{1+\lambda \epsilon}\E \left[(\dd f(x_t) - g_t)\tt  x_{t+1}\right] \\
    &\leq \frac{1}{1+\lambda\epsilon}\sqrt{\frac{d \cdot v_{max}}{n_{batch}}}
\ee

\end{proof}

\begin{lem}\label{lem:XY}
Let $X,Y$ be two $\RR^d$-valued random variables with $\var(Y)<+\infty$,  and assume $\phi$ yields a weak derivative $sign$. We have
$\E[(Y - \E Y)\tt sign(X+\epsilon Y)] \leq \sqrt{d \var(Y)}$
\end{lem}

\begin{proof} 
\bb 
\E[ (Y-\E [Y])\tt sign(X + \eps Y)] 
\leq \E [|Y-\E [Y]|] 
\leq \sqrt{d\cdot\E [\|Y-\E [Y]\|^2]} = \sqrt{d \cdot \var(Y)}
\ee 
\end{proof}

\end{document}